\pdfoutput=1
\documentclass[11pt]{article}

\usepackage[final]{acl}

\usepackage{times}
\usepackage{latexsym}

\usepackage[T1]{fontenc}
\usepackage[utf8]{inputenc}

\usepackage{microtype}

\usepackage{inconsolata}

\usepackage{graphicx}

\usepackage{xspace}
\usepackage{graphicx}
\usepackage{booktabs}
\usepackage{multirow}
\usepackage[noabbrev,capitalize]{cleveref}
\usepackage[separate-uncertainty=true,retain-zero-uncertainty=true]{siunitx}
\usepackage{subcaption}
\usepackage{enumitem}
\usepackage{soul}
\usepackage{wrapfig}
\usepackage{titletoc}
\usepackage{enumitem}
\usepackage{xcolor}
\usepackage{listings}
\usepackage{ulem}
\usepackage{framed}
\usepackage{mathrsfs}
\setlist{nosep}

\newcommand{\modelnamens}{ChemToolAgent}
\newcommand{\modelname}{\modelnamens\xspace}

\newcommand{\modelsnamens}{CTA}
\newcommand{\modelsname}{\modelsnamens\xspace}

\newcommand{\mmlucns}{MMLU-Chemistry}
\newcommand{\mmluc}{\mmlucns\xspace}

\newcommand{\sbcns}{SciBench-Chemistry}
\newcommand{\sbc}{\sbcns\xspace}

\newcommand{\gpqacns}{GPQA-Chemistry}
\newcommand{\gpqac}{\gpqacns\xspace}

\newcommand{\smolinstructns}{SMolInstruct}
\newcommand{\smolinstruct}{\smolinstructns\xspace}

\newcommand{\steptt}[1]{\noindent\textcolor{red}{\textbf{Step #1}\vspace{0.5em}}}
\newcommand{\sctt}[1]{\noindent\textcolor{blue}{{\small #1}\vspace{-0.5em}}}
\definecolor{codegreen}{rgb}{0,0.6,0}
\definecolor{codegray}{rgb}{0.5,0.5,0.5}
\definecolor{codepurple}{rgb}{0.58,0,0.82}
\definecolor{backcolour}{rgb}{0.95,0.95,0.92}
\definecolor{shadecolor}{rgb}{0.94,0.94,0.94}
\lstdefinestyle{mystyle}{
    backgroundcolor=\color{backcolour},   
    commentstyle=\color{codegreen},
    keywordstyle=\color{magenta},
    numberstyle=\tiny\color{codegray},
    stringstyle=\color{codepurple},
    basicstyle=\ttfamily\footnotesize,
    breakatwhitespace=false,         
    breaklines=true,                 
    captionpos=b,                    
    keepspaces=true,                 
    numbers=left,                    
    numbersep=5pt,                  
    showspaces=false,                
    showstringspaces=false,
    showtabs=false,                  
    tabsize=2
}
\newcommand{\newcontent}[1]{#1}

\title{\modelname: The Impact of Tools on Language Agents \\for Chemistry Problem Solving}

\author{
Botao Yu\textsuperscript{\dag$\mathcal{C}$}, Frazier N. Baker\textsuperscript{*$\mathcal{C}$$\mathcal{B}$}, Ziru Chen\textsuperscript{*$\mathcal{C}$}, Garrett Herb\textsuperscript{$\mathcal{C}$}, Boyu Gou\textsuperscript{$\mathcal{C}$},
\\
\textbf{Daniel Adu-Ampratwum\textsuperscript{$\mathcal{P}$}, Xia Ning\textsuperscript{$\mathcal{BCP}$}, Huan Sun\textsuperscript{\dag$\mathcal{C}$}}
\vspace{0.3em} \\
\small{\textsuperscript{$\mathcal{C}$}Department of Computer Science and Engineering, OSU} \\
\small{\textsuperscript{$\mathcal{B}$}Department of Biomedical Informatics, OSU} \qquad
\small{\textsuperscript{$\mathcal{P}$}College of Pharmacy, OSU} \\
\\
\small{Code: \url{https://github.com/OSU-NLP-Group/ChemToolAgent}} \\
\small{Toolkit: \url{https://osu-nlp-group.github.io/ChemMCP}}
}

\begin{document}

\maketitle
\def\thefootnote{*}\footnotetext{Equal contribution.}
\def\thefootnote{\textsuperscript{\dag}}\footnotetext{Correspondence to: \texttt{\{yu.3737, sun.397\}@osu.edu}.}
\def\thefootnote{\arabic{footnote}}
\begin{abstract}

To enhance large language models (LLMs) for chemistry problem solving, several LLM-based agents augmented with tools have been proposed, such as ChemCrow and Coscientist. However, their evaluations are narrow in scope, leaving a large gap in understanding the benefits of tools across diverse chemistry tasks. To bridge this gap, we develop ChemToolAgent, an enhanced chemistry agent over ChemCrow, and conduct a comprehensive evaluation of its performance on both specialized chemistry tasks and general chemistry questions. Surprisingly, ChemToolAgent does not consistently outperform its base LLMs without tools. Our error analysis with a chemistry expert suggests that: For specialized chemistry tasks, such as synthesis prediction, we should augment agents with specialized tools; however, for general chemistry questions like those in exams, agents' ability to reason correctly with chemistry knowledge matters more, and tool augmentation does not always help.

\end{abstract}

\section{Introduction}

Large language models (LLMs) have demonstrated impressive problem-solving capabilities in many disciplines \citep{wang2024mmlupro,yue2024mmmu,grossmann2023ai}. When it comes to chemistry, LLMs still face significant challenges, such as incorrect calculation, lack of domain knowledge, or inability to perform certain tasks like reaction prediction \citep{guo2023eighttask,mirza2024chembench}. To address these limitations, LLM-based agents integrated with tools have been proposed to tackle chemistry-specific problems \citep{wang2024agentsurvey,ramos2024chemistryagentsurvey}. For example, ChemCrow \citep{bran2023chemcrow} expands LLMs' capabilities by incorporating 18 tools, ranging from web search to chemical reaction prediction. Similarly, Coscientist \citep{boiko2023coscientist} integrates the control of cloud labs to enable LLMs to automate wet lab experiments.

Despite the promise of these tool-augmented agents, existing evaluations have been largely qualitative and limited in scope.
For example, ChemCrow is assessed with only 14 individual tasks mainly focusing on compound synthesis, and Coscientist's evaluation involves merely six specific tasks.
These narrow assessments leave a large gap in our understanding of how tool-augmented agents perform across diverse chemistry tasks in real-world applications.

In this work, we conduct a comprehensive evaluation of LLM-based agents on different chemistry tasks to grasp a deep understanding of their potential and limitations. 
To explore and enhance the capabilities of agents in diverse and complex chemistry scenarios, we introduce \newcontent{\modelname (\modelsname)}, a new chemistry agent capable of handling a wide spectrum of tasks. It leverages the ReAct framework \citep{yao2023react} and integrates 29 tools, such as a search tool for PubChem \citep{kim2019pubchem}, several molecular property predictors, as well as many practical tools present in ChemCrow. 
Then, we adapt two categories of real-world chemistry problems for systematic evaluation: specialized tasks and general questions.
For specialized tasks, we use \smolinstruct \citep{yu2024llasmol}, which contains 14 types of specialized molecule- and reaction-centric tasks. 
For general questions, we use \mmluc, \newcontent{\sbc}, and \gpqac, which are chemistry-related subsets of the MMLU \citep{hendryckstest2021mmlu}, \newcontent{SciBench \citep{wang2024scibench}}, and GPQA \citep{rein2023gpqa} benchmarks, containing exam-like questions spanning high school, college, and graduate levels.

Through comprehensive experiments, we show that: While \newcontent{\modelsname} substantially outperforms ChemCrow on all chemistry tasks, it does not consistently outperform the base LLMs without tools.
In addition, the impact of tool augmentation is highly dependent on task characteristics.
For specialized chemistry tasks involving professional molecular representations (e.g., SMILES \citep{weininger1988smiles}) and specialized chemical operations (e.g., compound synthesis), augmenting LLMs with task-specific tools can yield substantial performance gains. 
Nonetheless, for general chemistry questions that require fundamental knowledge and extensive reasoning, \newcontent{\modelsname} cannot address these challenges adequately and underperforms the base LLMs.
Further analysis along with a chemistry expert shows that \newcontent{\modelsname}'s underperformance on general chemistry questions is primarily due to nuanced mistakes at intermediate stages of its problem-solving process, such as wrong logic and information oversight.
Overall, our findings indicate that tool augmentation may introduce additional complexity that hinders LLMs' reasoning and thus does not always help in chemistry problem-solving.
Future research may improve LLM-based agents for chemistry by optimizing cognitive load and enhancing reasoning and information verification abilities.

\section{\modelname}
\label{sec:model}

\begin{figure}[t]
    \centering
    \includegraphics[width=0.9\columnwidth]{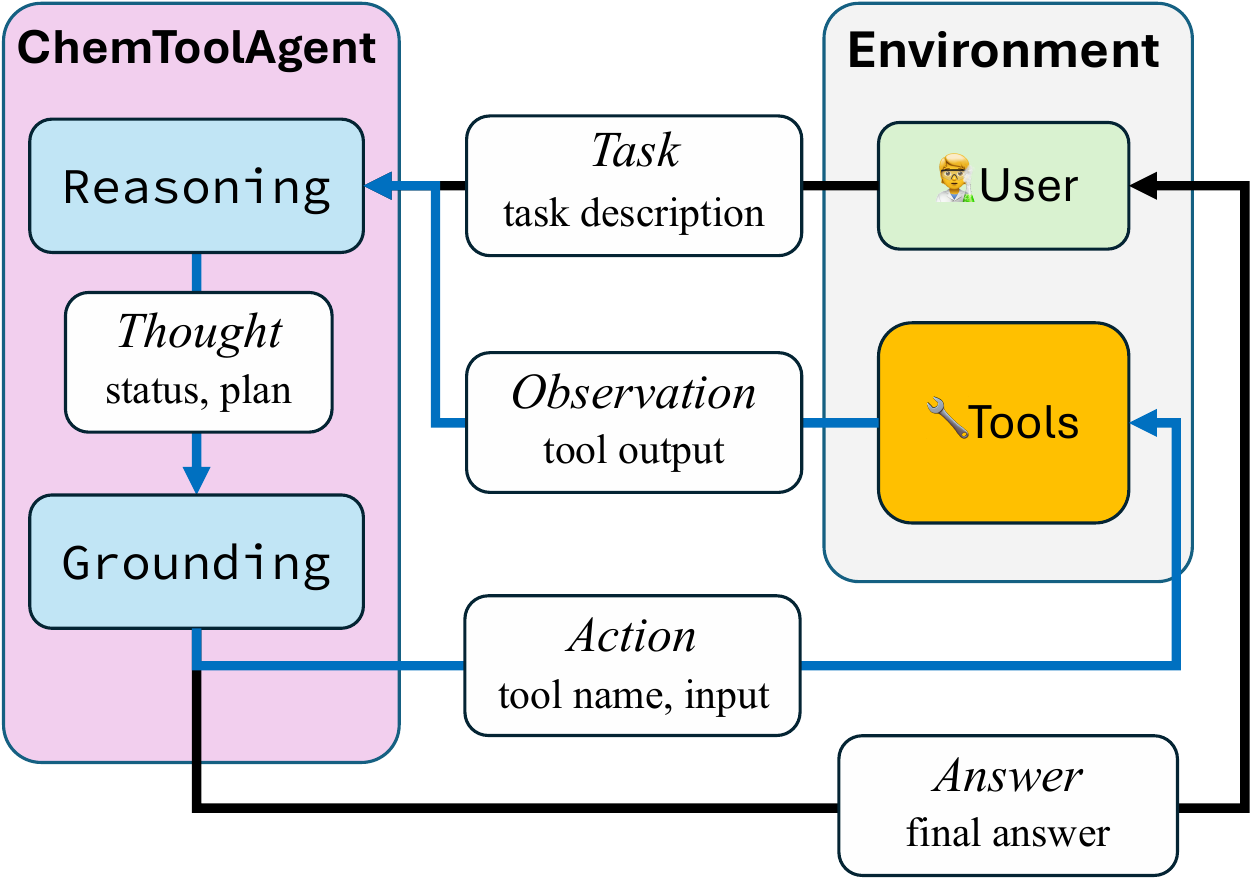}
    \caption{Our \modelname framework. Upon receiving a user \textit{task}, the agent iterates through a three-step ReAct process \citep{yao2023react}: (1) \textit{Thought} generation, analyzing the current situation and planning subsequent steps; (2) \textit{Action} determination, selecting the appropriate tool and its input based on the generated thought; and (3) \textit{Observation} obtaining, executing a tool in the environment and obtaining the results or feedback. This iterative cycle continues until task completion or conclusion, and the final \textit{answer} is returned to the user.}
    \label{fig:framework}
    % \vspace{-15pt}
    % }
\end{figure}

We introduce \modelname (\newcontent{or \modelsname for short; see its framework in \cref{fig:framework}}), a chemistry agent improved over ChemCrow \citep{bran2023chemcrow} and equipped with enhanced tools for a wider range of tasks.
It implements two essential cognitive abilities \citep{sumers2024cognitive} required for chemistry problem-solving:
(1) \textbf{Reasoning}: This ability is required in the \textit{Thought} step for comprehending user queries and tool outputs, assessing current status, and formulating subsequent steps. (2) \textbf{Grounding}: Based on the reasoning result (i.e., the ``thought''), this ability determines the appropriate tool to execute and its corresponding input. 

% \begin{wrapfigure}{r}{0.39\textwidth}
% \end{wrapfigure}

To enhance \newcontent{\modelsname}'s capabilities, we develop an extensive set of 29 tools (\cref{app:tool_set}), categorized into general, molecule, and reaction tools. 
General tools provide the agent with common problem-solving abilities, such as the execution of Python code for computations and various operations via \texttt{PythonREPL}.
Molecule tools specialize in the analysis, prediction, and conversion of molecules and their properties. 
For example, \texttt{FunctionalGroups} can identifies functional groups within a molecule, which is crucial for analyzing molecular characteristics. 
Lastly, reaction tools are instrumental in predicting chemical reaction outcomes (\texttt{ForwardSynthesis}) and suggesting synthesis paths for desired products (\texttt{Retrosynthesis}), both of which are essential in applications like drug discovery \citep{berdigaliyev2020overview}.

In this tool set, we create 16 new tools and enhance 6 existing ones in ChemCrow, which provides \newcontent{\modelsname} more comprehensive and robust abilities in solving chemistry problems. For example, we create \texttt{PubchemSearchQA}, which leverages an LLM to retrieve and extract authorized and comprehensive compound information from PubChem \citep{kim2019pubchem}, and several molecular property predictors (\texttt{BBBPPredictor}, \texttt{SideEffectPredictor}, etc.), which employ neural networks \citep{zhou2023unimol} for molecular property predictions. 
We also enhance \texttt{WebSearch} with an LLM-enhanced searching service to yield more comprehensive and flexible search results.

\section{Experiments}

\subsection{Experimental Setup}

\textbf{Datasets.} We use three well-established datasets (listed in \cref{tab:datasets}) to thoroughly assess tool-augmented agents on two categories of chemistry problems:
(1) \textbf{Specialized chemistry tasks} focus on experiment-like problems involving molecular manipulations, predictions, and representations. This category includes \textbf{\smolinstruct} \citep{yu2024llasmol}, which contains 14 molecule- and reaction-centric tasks and requires models to understand molecular representations like SMILES \citep{weininger1988smiles} and perform specific chemical operations, such as predicting synthesis paths and converting chemical names (\cref{fig:smolinstruct_tasks}).
(2) \textbf{General chemistry questions} resemble questions appearing in exams at different levels and test a wide range of fundamental knowledge and general reasoning in chemistry. This category includes \textbf{\mmluc}, a manually verified chemistry subset of the MMLU benchmark \citep{hendryckstest2021mmlu} that consists of questions at the high school and college level (\cref{app:mmluc}); \newcontent{\textbf{\sbc}, a set of the college-level chemistry calculation questions from SciBench \citep{wang2024scibench};} and \textbf{\gpqac}, the chemistry section of the GPQA-Diamond benchmark \citep{rein2023gpqa} that consists of difficult graduate-level questions.

% % Table generated by Excel2LaTeX from sheet 'Datasets'
% \begin{table}[htbp]
%   \centering
%   \caption{Datasets used in the evaluation.}
%     \scalebox{0.9}{
%     \begin{tabular}{llrl}
%     \toprule
%     \textbf{Dataset} & \textbf{Question format} & \textbf{\# Sample} & \textbf{Task type} \\
%     \midrule
%     SMolInstruct & Short answer & 700   & Specific molecule-centric tasks \\
%     MMLU-Chemistry & MCQ   & 70    & General questions \\
%     GPQA-Chemistry & MCQ   & 93    & Difficult general questions \\
%     ChemCrowTask & Open-ended question & 4     & Multifaceted practical tasks \\
%     \bottomrule
%     \end{tabular}%
%     }
%   \label{tab:datasets}%
% \end{table}%

% % Table generated by Excel2LaTeX from sheet 'Datasets'
% \begin{wraptable}{r}{0.6\textwidth}
\begin{table}[t]
  \centering
  \resizebox{\columnwidth}{!}{
    \begin{tabular}{p{1.8cm}p{3.3cm}rp{3.8cm}}
    \toprule
    \textbf{Category} & \textbf{Dataset} & \textbf{\# Sample} & \textbf{Specific task type} \\
    \midrule
    \multirow{1}[0]{=}{Specialized tasks} & SMolInstruct  & 700   & Molecule- and reaction-centric tasks \\
          % & ChemCrowTask & 4     & Multifaceted practical tasks \\
    \midrule
    \multirow{3}[2]{=}{General questions} & MMLU-Chemistry  & 70    & High school- and college-level questions \\
    \cmidrule(lr){2-4}
          & \newcontent{\sbc}  & \newcontent{223}    & \newcontent{College-level calculation questions} \\
    \cmidrule(lr){2-4}
          & GPQA-Chemistry  & 93    & Graduate-level questions \\
    \bottomrule
    \end{tabular}%
  }
  \caption{Datasets used in our experiments.}
  \label{tab:datasets}%
  % \vspace{-11pt}
\end{table}%
% \end{wraptable}

\noindent \textbf{LLMs and Agents.} We compare our \newcontent{\modelsname} with two baselines: (1)
State-of-the-art (SoTA) base LLMs, including GPT-4o \citep{gpt4o} and Claude-3.5-Sonnet \citep{claude3.5sonnet}, which have shown superior capabilities in chemistry problem-solving among existing LLMs \citep{wang2024mmlupro}. (2)
ChemCrow \citep{bran2023chemcrow}, a pioneering chemistry-focused agent equipped with 18 expert-designed tools. 
For ChemCrow and \newcontent{\modelsname}, we utilize GPT-4o or Claude-3.5-Sonnet as the backbone language models,
and refer to them as GPT and Claude, respectively.\footnote{gpt-4o-2024-08-06 and claude-3-5-sonnet-20240620.}

\subsection{Overall Performance}
\label{sec:exp_overall}

\noindent \textbf{Specialized Chemistry Tasks.}
Models are evaluated on 50 randomly selected samples from the test set of \smolinstruct for each task, and the results on four selected tasks are presented in \cref{tab:smolinstruct} (see \cref{app:detail_sm_results} for the full results).
We can observe that:
(1) \newcontent{\modelsname} exhibits substantial improvements over its base LLM counterparts,
highlighting the critical role of domain-specific tools in augmenting LLMs' capabilities on the specialized tasks in \smolinstruct. 
(2) Compared to ChemCrow, \newcontent{\modelsname} demonstrates superior performance. Our analysis suggests that the disparity is attributed to ChemCrow's limited tool set and the potential lack of robustness in its tool implementations. For instance, ChemCrow's apparent deficiency in molecular property prediction tools and its limited web search capabilities seem to hinder its performance in property prediction tasks. 
In contrast, \newcontent{\modelsname}'s tool set (\cref{app:tool_set}) is more comprehensive and robust for LLMs to leverage effectively.

% Table generated by Excel2LaTeX from sheet 'Overall_SMolInstruct'
\begin{table}[t]
  \centering
  \sisetup{detect-weight=true}
  % \begin{minipage}{0.45\textwidth}
  \resizebox{\columnwidth}{!}{
    \begin{tabular}{lS[table-format=2.1]S[table-format=2.1]S[table-format=2.1]S[table-format=2.1]}
    \toprule
    \multicolumn{1}{c}{\multirow{2}[2]{*}{\textbf{Model}}} & \multicolumn{1}{c}{\textbf{NC-S2I}} & \multicolumn{1}{c}{\textbf{PP-SIDER}} & \multicolumn{1}{c}{\textbf{FS}} & \multicolumn{1}{c}{\textbf{RS}} \\
          & \multicolumn{1}{c}{\textbf{EM}} & \multicolumn{1}{c}{\textbf{Acc}} & \multicolumn{1}{c}{\textbf{EM}} & \multicolumn{1}{c}{\textbf{EM}} \\
    \midrule
    % SoTA non-LLM models & $\underline{54.0}$  & $\underline{70.0}$  & $\underline{78.0}$  & \bfseries 42.0 \\
    GPT-4o & 0.0   & 44.0  & 12.0  & 0.0 \\
    Claude-3.5-Sonnet & 2.0   & 62.0  & 22.0  & 0.0 \\
    % LlaSMol & 34.0  & \bfseries 74.0  & 56.0  & $\underline{26.0}$ \\
    \midrule
    ChemCrow (GPT) & 2.0   & 36.0  & 72.0  & 8.0 \\
    ChemCrow (Claude) & 2.0   & 32.0  & 70.0  & 22.0 \\
    \midrule
    \modelsname (GPT) & \bfseries 70.0  & \bfseries 70.0  & 78.0  & \bfseries 42.0 \\
    \modelsname (Claude) & \bfseries 70.0  & 68.0  & \bfseries 80.0  & \bfseries 42.0 \\
    \bottomrule
    \end{tabular}%
    }
  \caption{The results (\%) on the \smolinstruct dataset. EM (exact match) and Acc (accuracy) are the metrics.}
  \label{tab:smolinstruct}%
  % \end{minipage}\hfill
\end{table}

\begin{table}[t]
  \centering
  \sisetup{detect-weight=true}
  % \begin{minipage}{0.53\textwidth}
  \resizebox{\columnwidth}{!}{
    \begin{tabular}{lS[table-format=2.1]S[table-format=2.1]S[table-format=2.1]}
    \toprule
    \textbf{Model} & \textbf{MMLU-C} & \textbf{SciBench-C} & \textbf{GPQA-C} \\
    \midrule
    GPT-4o & \bfseries 80.5 & 60.7 & 40.5 \\
    Claude-3.5-Sonnet & 76.7 & 60.7 & \bfseries  52.3 \\
    \midrule
    ChemCrow (GPT) & 43.3 & 52.5 & 27.5 \\
    ChemCrow (Claude) & 68.6 & 56.1 & 35.2 \\
    \midrule
    \modelsname (GPT) & 71.0 & 60.1 & 33.8 \\
    \modelsname (Claude) & 70.0 & \bfseries 65.9 & 45.9 \\
    \bottomrule
    \end{tabular}%
    }
  \caption{The accuracy scores (\%) on the \mmluc, \sbc, and \gpqac datasets, averaged over three runs.}
  \label{tab:mmluc_gpqac}%
  % \end{minipage}
  %\vspace{-11pt}
\end{table}%

\noindent \textbf{General Chemistry Questions.}
As presented in \cref{tab:mmluc_gpqac}, \newcontent{while \modelsname continues to outperforms ChemCrow (and also the most recent, concurrent ChemAgent \citep{tang2025chemagent}, see \cref{app:detail_sbc_results} for details), demonstrating its status as one of the top-performing chemistry agents to date, it surprisingly underperforms its base LLM counterparts in most cases, contrary to expectations}.
This trend persists across \newcontent{all three} datasets and is also observed with ChemCrow, suggesting a common issue in tool-augmented agents for chemistry. % consistent pattern rather than an isolated occurrence.
This observation challenges the intuition that tool augmentation would invariably enhance the performance of LLMs by providing additional information \citep{toolformer,qu2024tool}, and shows that both agents cannot fall back to base LLMs' capabilities when tools offer no advantage, calling for a second thought on applying such agents on different tasks with more thorough experiments.

\subsection{Error Analysis}
\label{sec:exp_error}

\newdimen\smheight
\settoheight{\smheight}{% <==========================================
  \includegraphics[width=0.30\textwidth]{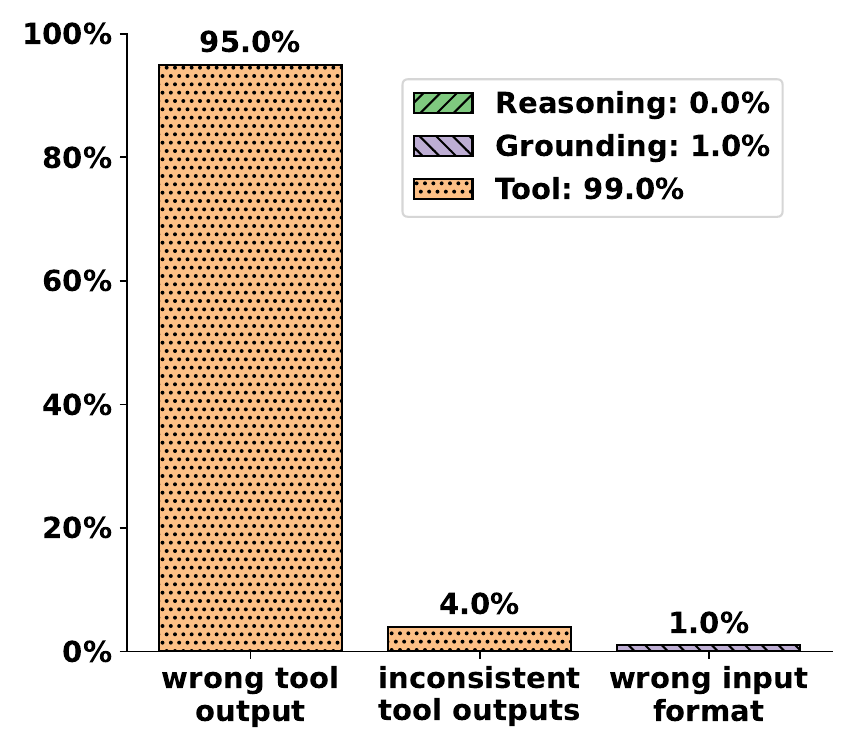}
}

\begin{figure*}[t]
    \centering
    \begin{subfigure}[b]{0.27\textwidth}
        \centering %,height=2in
        \includegraphics[height=\smheight]{fig/ChemAgent_GPT_SM_error_analysis.pdf}
        \caption{\smolinstruct.}
        \label{fig:ca_gpt_error_analysis_sm}
    \end{subfigure}
    \hfill
    \begin{subfigure}[b]{0.72\textwidth}
        \centering % , height=1.95in
        \includegraphics[height=\smheight]{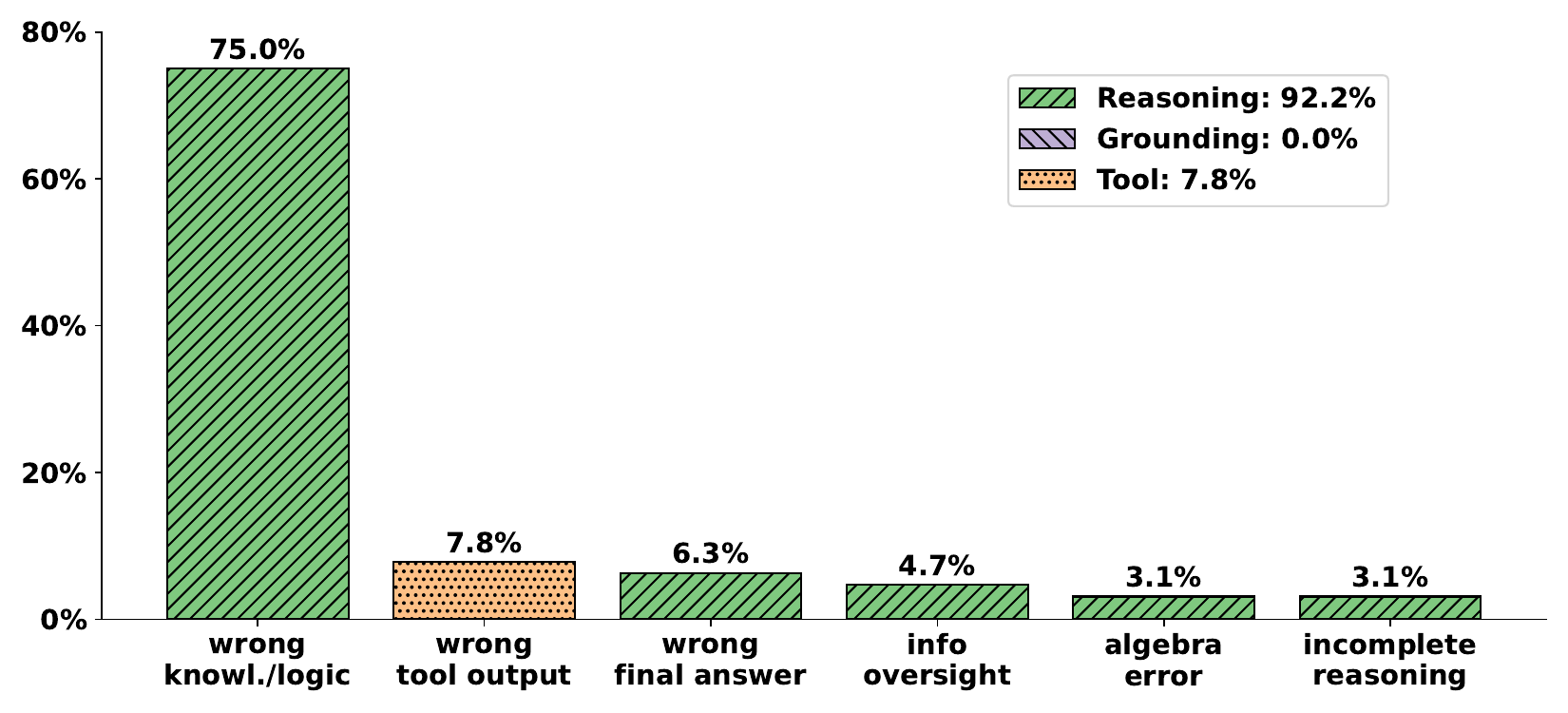}
        \caption{\mmluc.}
        \label{fig:ca_gpt_error_analysis_mmluc}
    \end{subfigure}
    \caption{The error statistics of \modelsname (GPT) on \smolinstruct (102 errors) and \mmluc (64 errors).}
    \label{fig:ca_gpt_error_analysis}
    \vspace{-11pt}
    % }
\end{figure*}

To examine the errors made by \newcontent{\modelsname}, we use \smolinstruct and \mmluc as representatives from their respective categories and conduct a manual error analysis. For all samples where \newcontent{\modelsname} (GPT) fails in our experiments, we engage a chemistry expert to analyze the errors, which are then classified into three types, namely \textbf{reasoning error}, \textbf{grounding error}, and \textbf{tool error}, based on the components (the cognitive abilities and the environment) responsible for the errors. The definitions of the identified errors identified are as follows:

\noindent \textbf{Reasoning errors.}
Errors made by the ``reasoning'' ability, where the agent inaccurately assesses the situation or devises an incorrect plan for subsequent steps, such as misinterpreting tool outputs or suggesting incorrect methodologies. Specifically, they include the following errors:
\begin{itemize}[leftmargin=*]
\item Wrong knowledge/logic: an error where agent makes a mistake in applying chemistry knowledge or makes a conclusion that does not logically follow from the previous information. 
\item Wrong final answer: an error where the analysis process is correct but the final answer is wrong.
\item Information oversight: an error where the agent neglects to consider relevant information given in the question or the previous steps.
\item Algebra error: an error in algebraic manipulation or simplification, such as the incorrect solving of equations or misapplication of algebraic axioms. % , e.g. substitution or division properties of equality.
\item Incomplete reasoning: An error where the reasoning process is not fully developed, such as 
when solving a problem but omitting necessary steps or details. 
\end{itemize}

\noindent \textbf{Grounding errors.}
These occur during tool invocation, such as selecting an inappropriate tool, using an incorrect input format, or providing erroneous inputs to a tool. Specifically:
\begin{itemize}[leftmargin=*]
\item Wrong input format: an error arising from data being provided in a format that the tool cannot process, resulting in failures or incorrect results.
\end{itemize}

\noindent \textbf{Tool errors.}
These errors originate from the environment (i.e., the tools used in this study), where the tools either fail to execute properly or return inaccurate information. Specifically:
\begin{itemize}[leftmargin=*]
\item Wrong tool output: an error occurring when a tool produces incorrect or unexpected results, leading to faulty conclusions or actions.
\item Inconsistent tool outputs: an error where multiple tools return inconsistent information, leading to faulty conclusions or actions.
\end{itemize}

As illustrated in \cref{fig:ca_gpt_error_analysis}, the error distributions are very different on the two datasets. 
On \smolinstruct (\cref{fig:ca_gpt_error_analysis_sm}), tool errors account for over 99.0\% of all errors.
These errors mainly stem from the neural networks-based tools (e.g., \texttt{ForwardSynthesis}, \texttt{BBBPPredictor}), which inherently possess imperfect accuracy.
For these specialized tasks where dedicated tools exist, the agent can easily pinpoint and correctly use the needed tools (\cref{app:tool_analysis}), resulting in limited or no reasoning and grounding errors.
In contrast, on \mmluc (\cref{fig:ca_gpt_error_analysis_mmluc}), reasoning errors constitute over 90.0\%. This is because MMLU questions require broader knowledge and more intricate chemical reasoning and rely less on external tools.
Our analysis indicates that all the observed reasoning errors manifest as nuanced mistakes at intermediate stages of problem-solving, rather than incorrect overall methods. 
For instance, an inaccurate chemistry knowledge is applied or a mistaken conclusion is made (the wrong knowledge/logic error), or a wrong final option is selected despite of the correct analysis process (the wrong final answer error).
Specific cases showcasing the errors can be found in \cref{app:errors}.
Compared to LLMs without tools, the tool-augmented agent appears more prone to such nuanced mistakes.

We hypothesize that the errors occur for two reasons:
(1) \textbf{Increased cognitive load of agents}: The backbone LLM in the agent is tasked with multiple responsibilities, including task comprehension, tool selection, and tool output interpretation. This necessitates frequent role-switching \citep{qiao2024autoact}, which, along with more complex contexts, may hinder the LLM' ability to maintain a holistic and consistent approach to the main task \citep{verma2024brittle}, resulting in more reasoning errors.
(2) \textbf{Potentially confusing tool outputs}: The tool outputs may occasionally be inaccurate or conflict with the model's internal knowledge \citep{xie2024adaptive}. This discrepancy can introduce confusion and lead to reasoning and tool errors.
To address these issues, future research could focus on developing new agent frameworks that \textbf{reduce cognitive load and context distractions for LLMs}. This might involve \textit{building multi-agent systems} to distribute the workload \citep{chen2023reconcile,chen2024magdi}, or \textit{filtering out irrelevant information} to enhance task-focused reasoning \citep{shi2023distracted,yuan2024easytool,ouyang2023structured}. Additionally, exploring \textbf{information verification mechanisms} could help LLMs resolve discrepancies from multiple sources, improving the accuracy of the final output.

\section{Conclusion}

In this paper, we conducted a comprehensive evaluation of tool-augmented language agents for chemistry problem-solving. We introduce \modelname, an enhanced chemistry agent with an enhanced tool set, and assess its performance across diverse chemistry problems, including specialized tasks and general questions.
Our findings reveal that the impact of tool augmentation is highly dependent on task characteristics: While \modelname demonstrates significant improvements on specialized tasks, it does not surpass the base LLMs without tools on general questions.
The manual error analysis highlights that tool errors predominate in specialized tasks, whereas reasoning errors are more frequent in general questions due to nuanced mistakes in the problem solving process. To minimize reasoning errors and enhance performance in general questions, future agent design should focus on optimizing the cognitive load of LLMs and improving their ability to reason and verify information, especially when resolving inconsistencies from multiple sources.

\section*{Limitations}

This study focuses on evaluating the performance of tool-augmented agents across various chemistry tasks. While it presents the most comprehensive evaluation on this topic to the best of our knowledge, there are several limitations:

\begin{itemize}[leftmargin=*]
    \item Our evaluation uses GPT-4o and Claude-3.5-Sonnet as the primary models for comparison and as backbones for the agents. This selection does not encompass a broader range of potential LLMs, such as Llama-3.2 \citep{llama3.2} or Qwen2 \citep{yang2024qwen2}. Although these additional models might exhibit different performance patterns, our choice of state-of-the-art (SoTA) models ensures strong baselines  without significant loss of generality.
    \item While our study addresses both specialized tasks and general questions in the field of chemistry, it may not entirely represent the vast array of real-world chemistry problems. The scope might overlook certain nuanced challenges and scenarios encountered in practical applications.
    \item The manual error analysis, while thorough, is conducted by a chemistry expert together with two PhD students with chemistry knowledge and is limited in scale. This constraint could introduce potential biases and may not capture all subtle variations in error types or frequencies.
\end{itemize}

\section*{Acknowledgments}
The authors would thank colleagues from the OSU NLP group and the OSU Ning Lab for their constructive feedback. This research was supported in part by NSF CAREER \#1942980, NLM R01LM014385, and NSF IIS-2133650. The views and conclusions contained herein are those of the authors and should not be interpreted as representing the official policies, either expressed or implied, of the U.S. government. The U.S. Government is authorized to reproduce and distribute reprints for Government purposes notwithstanding any copyright notice herein.

\bibliography{reference}

\begin{thebibliography}{54}
\providecommand{\natexlab}[1]{#1}

\bibitem[{Anthropic(2024)}]{claude3.5sonnet}
Anthropic. 2024.
\newblock Introducing claude 3.5 sonnet.
\newblock \url{https://www.anthropic.com/news/claude-3-5-sonnet}.
\newblock Accessed: 2024-10-15.

\bibitem[{Berdigaliyev and Aljofan(2020)}]{berdigaliyev2020overview}
Nurken Berdigaliyev and Mohamad Aljofan. 2020.
\newblock An overview of drug discovery and development.
\newblock \emph{Future medicinal chemistry}, 12(10):939--947.

\bibitem[{Boiko et~al.(2023)Boiko, MacKnight, Kline, and Gomes}]{boiko2023coscientist}
Daniil~A Boiko, Robert MacKnight, Ben Kline, and Gabe Gomes. 2023.
\newblock Autonomous chemical research with large language models.
\newblock \emph{Nature}, 624(7992):570--578.

\bibitem[{Chen et~al.(2023)Chen, Saha, and Bansal}]{chen2023reconcile}
Justin Chih-Yao Chen, Swarnadeep Saha, and Mohit Bansal. 2023.
\newblock Reconcile: Round-table conference improves reasoning via consensus among diverse llms.
\newblock \emph{arXiv preprint arXiv:2309.13007}.

\bibitem[{Chen et~al.(2024{\natexlab{a}})Chen, Saha, Stengel-Eskin, and Bansal}]{chen2024magdi}
Justin Chih-Yao Chen, Swarnadeep Saha, Elias Stengel-Eskin, and Mohit Bansal. 2024{\natexlab{a}}.
\newblock Magdi: Structured distillation of multi-agent interaction graphs improves reasoning in smaller language models.
\newblock \emph{arXiv preprint arXiv:2402.01620}.

\bibitem[{Chen et~al.(2024{\natexlab{b}})Chen, Li, Wang, Du, Yu, Lu, Li, Qiu, Pan, Huang, Fang, Heng, and Chen}]{chen2024chemistx}
Kexin Chen, Junyou Li, Kunyi Wang, Yuyang Du, Jiahui Yu, Jiamin Lu, Lanqing Li, Jiezhong Qiu, Jianzhang Pan, Yi~Huang, Qun Fang, Pheng~Ann Heng, and Guangyong Chen. 2024{\natexlab{b}}.
\newblock Chemist-x: Large language model-empowered agent for reaction condition recommendation in chemical synthesis.
\newblock \emph{arXiv preprint arXiv:2311.10776}.

\bibitem[{Chen et~al.(2024{\natexlab{c}})Chen, Lin, Sch{\"a}rli, and Zhou}]{chen2024teaching}
Xinyun Chen, Maxwell Lin, Nathanael Sch{\"a}rli, and Denny Zhou. 2024{\natexlab{c}}.
\newblock Teaching large language models to self-debug.
\newblock In \emph{Proceedings of International Conference on Learning Representations (ICLR)}.

\bibitem[{Chen et~al.(2024{\natexlab{d}})Chen, Chen, Ning, Zhang, Wang, Yu, Li, Liao, Wei, Lu et~al.}]{chen2024scienceagentbench}
Ziru Chen, Shijie Chen, Yuting Ning, Qianheng Zhang, Boshi Wang, Botao Yu, Yifei Li, Zeyi Liao, Chen Wei, Zitong Lu, et~al. 2024{\natexlab{d}}.
\newblock Scienceagentbench: Toward rigorous assessment of language agents for data-driven scientific discovery.
\newblock \emph{arXiv preprint arXiv:2410.05080}.

\bibitem[{CHERUKARA et~al.(2024)CHERUKARA, VRIZA, CHAN, ZHOU, KATTI~SASTRY, and PRINCE}]{cherukara2024calms}
MATTHEW CHERUKARA, ALKATERINI VRIZA, HENRY CHAN, TAO ZHOU, VARUNI KATTI~SASTRY, and MICHAEL PRINCE. 2024.
\newblock Calms: Context-aware language model for science.
\newblock Technical report, Argonne National Laboratory (ANL), Argonne, IL (United States).

\bibitem[{Edwards et~al.(2022)Edwards, Lai, Ros, Honke, Cho, and Ji}]{edwards2022translation}
Carl Edwards, Tuan Lai, Kevin Ros, Garrett Honke, Kyunghyun Cho, and Heng Ji. 2022.
\newblock Translation between molecules and natural language.
\newblock In \emph{Proceedings of Conference on Empirical Methods in Natural Language Processing (EMNLP)}, pages 375--413.

\bibitem[{Gema et~al.(2024)Gema, Leang, Hong, Devoto, Mancino, Saxena, He, Zhao, Du, Madani et~al.}]{gema2024mmluredux}
Aryo~Pradipta Gema, Joshua Ong~Jun Leang, Giwon Hong, Alessio Devoto, Alberto Carlo~Maria Mancino, Rohit Saxena, Xuanli He, Yu~Zhao, Xiaotang Du, Mohammad Reza~Ghasemi Madani, et~al. 2024.
\newblock Are we done with mmlu?
\newblock \emph{arXiv preprint arXiv:2406.04127}.

\bibitem[{Ghafarollahi and Buehler(2024)}]{ghafarollahi2024protagents}
Alireza Ghafarollahi and Markus~J Buehler. 2024.
\newblock Protagents: protein discovery via large language model multi-agent collaborations combining physics and machine learning.
\newblock \emph{Digital Discovery}.

\bibitem[{Grossmann et~al.(2023)Grossmann, Feinberg, Parker, Christakis, Tetlock, and Cunningham}]{grossmann2023ai}
Igor Grossmann, Matthew Feinberg, Dawn~C Parker, Nicholas~A Christakis, Philip~E Tetlock, and William~A Cunningham. 2023.
\newblock Ai and the transformation of social science research.
\newblock \emph{Science}, 380(6650):1108--1109.

\bibitem[{Guo et~al.(2023)Guo, Guo, Nan, Liang, Guo, Chawla, Wiest, and Zhang}]{guo2023eighttask}
Taicheng Guo, Kehan Guo, Bozhao Nan, Zhenwen Liang, Zhichun Guo, Nitesh~V Chawla, Olaf Wiest, and Xiangliang Zhang. 2023.
\newblock What can large language models do in chemistry? a comprehensive benchmark on eight tasks.
\newblock In \emph{Proceedings of Neural Information Processing Systems (NeurIPS) Datasets and Benchmarks Track}.

\bibitem[{Hendrycks et~al.(2021)Hendrycks, Burns, Basart, Zou, Mazeika, Song, and Steinhardt}]{hendryckstest2021mmlu}
Dan Hendrycks, Collin Burns, Steven Basart, Andy Zou, Mantas Mazeika, Dawn Song, and Jacob Steinhardt. 2021.
\newblock Measuring massive multitask language understanding.
\newblock In \emph{Proceedings of International Conference on Learning Representations (ICLR)}.

\bibitem[{Huang et~al.(2024)Huang, Qu, Cousins, Johnson, Yin, Shah, Zhou, Altman, Wang, and Cong}]{huang2024crispr}
Kaixuan Huang, Yuanhao Qu, Henry Cousins, William~A Johnson, Di~Yin, Mihir Shah, Denny Zhou, Russ Altman, Mengdi Wang, and Le~Cong. 2024.
\newblock Crispr-gpt: An llm agent for automated design of gene-editing experiments.
\newblock \emph{arXiv preprint arXiv:2404.18021}.

\bibitem[{Jiang et~al.(2023)Jiang, Sablayrolles, Mensch, Bamford, Chaplot, de~las Casas, Bressand, Lengyel, Lample, Saulnier et~al.}]{jiang2023mistral}
Albert~Q Jiang, Alexandre Sablayrolles, Arthur Mensch, Chris Bamford, Devendra~Singh Chaplot, Diego de~las Casas, Florian Bressand, Gianna Lengyel, Guillaume Lample, Lucile Saulnier, et~al. 2023.
\newblock Mistral 7b.
\newblock \emph{arXiv preprint arXiv:2310.06825}.

\bibitem[{Kang and Kim(2023)}]{kang2023chatmof}
Yeonghun Kang and Jihan Kim. 2023.
\newblock Chatmof: An autonomous ai system for predicting and generating metal-organic frameworks.
\newblock \emph{arXiv preprint arXiv:2308.01423}.

\bibitem[{Kim et~al.(2019)Kim, Chen, Cheng, Gindulyte, He, He, Li, Shoemaker, Thiessen, Yu et~al.}]{kim2019pubchem}
Sunghwan Kim, Jie Chen, Tiejun Cheng, Asta Gindulyte, Jia He, Siqian He, Qingliang Li, Benjamin~A Shoemaker, Paul~A Thiessen, Bo~Yu, et~al. 2019.
\newblock Pubchem 2019 update: improved access to chemical data.
\newblock \emph{Nucleic acids research}, 47:D1102--D1109.

\bibitem[{Krenn et~al.(2019)Krenn, H{\"a}se, Nigam, Friederich, and Aspuru-Guzik}]{krenn2019selfies}
Mario Krenn, Florian H{\"a}se, A~Nigam, Pascal Friederich, and Al{\'a}n Aspuru-Guzik. 2019.
\newblock Selfies: a robust representation of semantically constrained graphs with an example application in chemistry.
\newblock \emph{arXiv preprint arXiv:1905.13741}, 1(3).

\bibitem[{M.~Bran et~al.(2024)M.~Bran, Cox, Schilter, Baldassari, White, and Schwaller}]{bran2023chemcrow}
Andres M.~Bran, Sam Cox, Oliver Schilter, Carlo Baldassari, Andrew~D White, and Philippe Schwaller. 2024.
\newblock Augmenting large language models with chemistry tools.
\newblock \emph{Nature Machine Intelligence}, pages 1--11.

\bibitem[{McNaughton et~al.(2024)McNaughton, Ramalaxmi, Kruel, Knutson, Varikoti, and Kumar}]{mcnaughton2024cactus}
Andrew~D McNaughton, Gautham Ramalaxmi, Agustin Kruel, Carter~R Knutson, Rohith~A Varikoti, and Neeraj Kumar. 2024.
\newblock Cactus: Chemistry agent connecting tool-usage to science.
\newblock \emph{arXiv preprint arXiv:2405.00972}.

\bibitem[{Meta(2024)}]{llama3.2}
Meta. 2024.
\newblock Llama 3.2.
\newblock \url{https://www.llama.com/docs/model-cards-and-prompt-formats/llama3_2}.
\newblock Accessed: 2024-10-15.

\bibitem[{Mirza et~al.(2024)Mirza, Alampara, Kunchapu, Emoekabu, Krishnan, Wilhelmi, Okereke, Eberhardt, Elahi, Greiner et~al.}]{mirza2024chembench}
Adrian Mirza, Nawaf Alampara, Sreekanth Kunchapu, Benedict Emoekabu, Aswanth Krishnan, Mara Wilhelmi, Macjonathan Okereke, Juliane Eberhardt, Amir~Mohammad Elahi, Maximilian Greiner, et~al. 2024.
\newblock Are large language models superhuman chemists?
\newblock \emph{arXiv preprint arXiv:2404.01475}.

\bibitem[{O'Donoghue et~al.(2023)O'Donoghue, Shtedritski, Ginger, Abboud, Ghareeb, Booth, and Rodriques}]{o2023bioplanner}
Odhran O'Donoghue, Aleksandar Shtedritski, John Ginger, Ralph Abboud, Ali~Essa Ghareeb, Justin Booth, and Samuel~G Rodriques. 2023.
\newblock Bioplanner: automatic evaluation of llms on protocol planning in biology.
\newblock \emph{arXiv preprint arXiv:2310.10632}.

\bibitem[{OpenAI(2023)}]{achiam2023gpt4}
OpenAI. 2023.
\newblock Gpt-4 technical report.
\newblock \emph{arXiv preprint arXiv:2303.08774}.

\bibitem[{OpenAI(2024)}]{gpt4o}
OpenAI. 2024.
\newblock Hello, gpt-4o.
\newblock \url{https://openai.com/index/hello-gpt-4o/}.
\newblock Accessed: 2024-10-15.

\bibitem[{Ouyang et~al.(2023)Ouyang, Zhang, Yan, Liu, Han, and Qin}]{ouyang2023structured}
Siru Ouyang, Zhuosheng Zhang, Bing Yan, Xuan Liu, Jiawei Han, and Lianhui Qin. 2023.
\newblock Structured chemistry reasoning with large language models.
\newblock \emph{arXiv preprint arXiv:2311.09656}.

\bibitem[{Qiao et~al.(2024)Qiao, Zhang, Fang, Luo, Zhou, Jiang, chengfei lv, and Chen}]{qiao2024autoact}
Shuofei Qiao, Ningyu Zhang, Runnan Fang, Yujie Luo, Wangchunshu Zhou, Yuchen~Eleanor Jiang, chengfei lv, and Huajun Chen. 2024.
\newblock Autoact: Automatic agent learning from scratch via self-planning.
\newblock In \emph{ICLR Workshop on Large Language Model (LLM) Agents}.

\bibitem[{Qu et~al.(2024)Qu, Dai, Wei, Cai, Wang, Yin, Xu, and Wen}]{qu2024tool}
Changle Qu, Sunhao Dai, Xiaochi Wei, Hengyi Cai, Shuaiqiang Wang, Dawei Yin, Jun Xu, and Ji-Rong Wen. 2024.
\newblock Tool learning with large language models: A survey.
\newblock \emph{arXiv preprint arXiv:2405.17935}.

\bibitem[{Rajan et~al.(2021)Rajan, Zielesny, and Steinbeck}]{rajan2021stout}
Kohulan Rajan, Achim Zielesny, and Christoph Steinbeck. 2021.
\newblock Stout: Smiles to iupac names using neural machine translation.
\newblock \emph{Journal of Cheminformatics}, 13(1):1--14.

\bibitem[{Ramos et~al.(2024)Ramos, Collison, and White}]{ramos2024chemistryagentsurvey}
Mayk~Caldas Ramos, Christopher~J Collison, and Andrew~D White. 2024.
\newblock A review of large language models and autonomous agents in chemistry.
\newblock \emph{arXiv preprint arXiv:2407.01603}.

\bibitem[{RDKit(2023)}]{rdkit}
RDKit. 2023.
\newblock \href {https://doi.org/10.5281/zenodo.8254217} {Rdkit: Open-source cheminformatics}.
\newblock Accessed on 27 Jan 2024.

\bibitem[{Rein et~al.(2023)Rein, Hou, Stickland, Petty, Pang, Dirani, Michael, and Bowman}]{rein2023gpqa}
David Rein, Betty~Li Hou, Asa~Cooper Stickland, Jackson Petty, Richard~Yuanzhe Pang, Julien Dirani, Julian Michael, and Samuel~R Bowman. 2023.
\newblock Gpqa: A graduate-level google-proof q\&a benchmark.
\newblock \emph{arXiv preprint arXiv:2311.12022}.

\bibitem[{Ruan et~al.(2024)Ruan, Lu, Xu, Zhang, Xuan, Pan, Fang, Gao, Shen, Ye et~al.}]{ruan2024accelerated}
Yixiang Ruan, Chenyin Lu, Ning Xu, Jian Zhang, Jun Xuan, Jianzhang Pan, Qun Fang, Hanyu Gao, Xiaodong Shen, Ning Ye, et~al. 2024.
\newblock Accelerated end-to-end chemical synthesis development with large language models.
\newblock \emph{ChemRxiv preprint}.

\bibitem[{Schick et~al.(2023)Schick, Dwivedi-Yu, Dessi, Raileanu, Lomeli, Hambro, Zettlemoyer, Cancedda, and Scialom}]{toolformer}
Timo Schick, Jane Dwivedi-Yu, Roberto Dessi, Roberta Raileanu, Maria Lomeli, Eric Hambro, Luke Zettlemoyer, Nicola Cancedda, and Thomas Scialom. 2023.
\newblock Toolformer: Language models can teach themselves to use tools.
\newblock In \emph{Proceedings of Neural Information Processing Systems (NeurIPS)}, volume~36, pages 68539--68551.

\bibitem[{Schwaller et~al.(2019)Schwaller, Laino, Gaudin, Bolgar, Hunter, Bekas, and Lee}]{schwaller2019molecular}
Philippe Schwaller, Teodoro Laino, Th{\'e}ophile Gaudin, Peter Bolgar, Christopher~A Hunter, Costas Bekas, and Alpha~A Lee. 2019.
\newblock Molecular transformer: a model for uncertainty-calibrated chemical reaction prediction.
\newblock \emph{ACS central science}, 5(9):1572--1583.

\bibitem[{Shi et~al.(2023)Shi, Chen, Misra, Scales, Dohan, Chi, Sch{\"a}rli, and Zhou}]{shi2023distracted}
Freda Shi, Xinyun Chen, Kanishka Misra, Nathan Scales, David Dohan, Ed~H Chi, Nathanael Sch{\"a}rli, and Denny Zhou. 2023.
\newblock Large language models can be easily distracted by irrelevant context.
\newblock In \emph{Proceedings of International Conference on Machine Learning (ICML)}, pages 31210--31227.

\bibitem[{Sumers et~al.(2024)Sumers, Yao, Narasimhan, and Griffiths}]{sumers2024cognitive}
Theodore Sumers, Shunyu Yao, Karthik Narasimhan, and Thomas Griffiths. 2024.
\newblock Cognitive architectures for language agents.
\newblock \emph{Transactions on Machine Learning Research}.

\bibitem[{Tang et~al.(2025)Tang, Hu, Ye, Shao, Yin, Ouyang, Zhou, Lu, Zhang, Zhao et~al.}]{tang2025chemagent}
Xiangru Tang, Tianyu Hu, Muyang Ye, Yanjun Shao, Xunjian Yin, Siru Ouyang, Wangchunshu Zhou, Pan Lu, Zhuosheng Zhang, Yilun Zhao, et~al. 2025.
\newblock Chemagent: Self-updating library in large language models improves chemical reasoning.
\newblock In \emph{Proceedings of International Conference on Learning Representations (ICLR)}.

\bibitem[{Verma et~al.(2024)Verma, Bhambri, and Kambhampati}]{verma2024brittle}
Mudit Verma, Siddhant Bhambri, and Subbarao Kambhampati. 2024.
\newblock On the brittle foundations of react prompting for agentic large language models.
\newblock \emph{arXiv preprint arXiv:2405.13966}.

\bibitem[{Wang et~al.(2022)Wang, Deng, and Sun}]{wang2022iteratively}
Boshi Wang, Xiang Deng, and Huan Sun. 2022.
\newblock Iteratively prompt pre-trained language models for chain of thought.
\newblock In \emph{Proceedings of Conference on Empirical Methods in Natural Language Processing (EMNLP)}, pages 2714--2730.

\bibitem[{Wang et~al.(2024{\natexlab{a}})Wang, Ma, Feng, Zhang, Yang, Zhang, Chen, Tang, Chen, Lin et~al.}]{wang2024agentsurvey}
Lei Wang, Chen Ma, Xueyang Feng, Zeyu Zhang, Hao Yang, Jingsen Zhang, Zhiyuan Chen, Jiakai Tang, Xu~Chen, Yankai Lin, et~al. 2024{\natexlab{a}}.
\newblock A survey on large language model based autonomous agents.
\newblock \emph{Frontiers of Computer Science}, 18(6):186345.

\bibitem[{Wang et~al.(2024{\natexlab{b}})Wang, Hu, Lu, Zhu, Zhang, Subramaniam, Loomba, Zhang, Sun, and Wang}]{wang2024scibench}
Xiaoxuan Wang, Ziniu Hu, Pan Lu, Yanqiao Zhu, Jieyu Zhang, Satyen Subramaniam, Arjun~R Loomba, Shichang Zhang, Yizhou Sun, and Wei Wang. 2024{\natexlab{b}}.
\newblock Scibench: Evaluating college-level scientific problem-solving abilities of large language models.
\newblock In \emph{Proceedings of International Conference on Machine Learning (ICML)}.

\bibitem[{Wang et~al.(2024{\natexlab{c}})Wang, Ma, Zhang, Ni, Chandra, Guo, Ren, Arulraj, He, Jiang et~al.}]{wang2024mmlupro}
Yubo Wang, Xueguang Ma, Ge~Zhang, Yuansheng Ni, Abhranil Chandra, Shiguang Guo, Weiming Ren, Aaran Arulraj, Xuan He, Ziyan Jiang, et~al. 2024{\natexlab{c}}.
\newblock Mmlu-pro: A more robust and challenging multi-task language understanding benchmark.
\newblock \emph{arXiv preprint arXiv:2406.01574}.

\bibitem[{Wei et~al.(2022)Wei, Wang, Schuurmans, Bosma, Xia, Chi, Le, Zhou et~al.}]{wei2022chain}
Jason Wei, Xuezhi Wang, Dale Schuurmans, Maarten Bosma, Fei Xia, Ed~Chi, Quoc~V Le, Denny Zhou, et~al. 2022.
\newblock Chain-of-thought prompting elicits reasoning in large language models.
\newblock \emph{Proceedings of Neural Information Processing Systems (NeurIPS)}, 35:24824--24837.

\bibitem[{Weininger(1988)}]{weininger1988smiles}
David Weininger. 1988.
\newblock Smiles, a chemical language and information system. 1. introduction to methodology and encoding rules.
\newblock \emph{Journal of chemical information and computer sciences}, 28(1):31--36.

\bibitem[{Xie et~al.(2024)Xie, Zhang, Chen, Lou, and Su}]{xie2024adaptive}
Jian Xie, Kai Zhang, Jiangjie Chen, Renze Lou, and Yu~Su. 2024.
\newblock Adaptive chameleon or stubborn sloth: Revealing the behavior of large language models in knowledge conflicts.
\newblock In \emph{The Twelfth International Conference on Learning Representations}.

\bibitem[{Yang et~al.(2024)Yang, Yang, Hui, Zheng, Yu, Zhou, Li, Li, Liu, Huang et~al.}]{yang2024qwen2}
An~Yang, Baosong Yang, Binyuan Hui, Bo~Zheng, Bowen Yu, Chang Zhou, Chengpeng Li, Chengyuan Li, Dayiheng Liu, Fei Huang, et~al. 2024.
\newblock Qwen2 technical report.
\newblock \emph{arXiv preprint arXiv:2407.10671}.

\bibitem[{Yao et~al.(2023)Yao, Zhao, Yu, Du, Shafran, Narasimhan, and Cao}]{yao2023react}
Shunyu Yao, Jeffrey Zhao, Dian Yu, Nan Du, Izhak Shafran, Karthik~R Narasimhan, and Yuan Cao. 2023.
\newblock React: Synergizing reasoning and acting in language models.
\newblock In \emph{Proceedings of International Conference on Learning Representations (ICLR)}.

\bibitem[{Yu et~al.(2024)Yu, Baker, Chen, Ning, and Sun}]{yu2024llasmol}
Botao Yu, Frazier~N. Baker, Ziqi Chen, Xia Ning, and Huan Sun. 2024.
\newblock Lla{SM}ol: Advancing large language models for chemistry with a large-scale, comprehensive, high-quality instruction tuning dataset.
\newblock In \emph{Proceedings of Conference on Language Modeling (COLM)}.

\bibitem[{Yuan et~al.(2024)Yuan, Song, Chen, Tan, Shen, Ren, Li, and Yang}]{yuan2024easytool}
Siyu Yuan, Kaitao Song, Jiangjie Chen, Xu~Tan, Yongliang Shen, Kan Ren, Dongsheng Li, and Deqing Yang. 2024.
\newblock {EASYTOOL}: Enhancing {LLM}-based agents with concise tool instruction.
\newblock In \emph{Proceedings of ICLR Workshop on Large Language Model (LLM) Agents}.

\bibitem[{Yue et~al.(2024)Yue, Ni, Zhang, Zheng, Liu, Zhang, Stevens, Jiang, Ren, Sun et~al.}]{yue2024mmmu}
Xiang Yue, Yuansheng Ni, Kai Zhang, Tianyu Zheng, Ruoqi Liu, Ge~Zhang, Samuel Stevens, Dongfu Jiang, Weiming Ren, Yuxuan Sun, et~al. 2024.
\newblock Mmmu: A massive multi-discipline multimodal understanding and reasoning benchmark for expert agi.
\newblock In \emph{Proceedings of the IEEE/CVF Conference on Computer Vision and Pattern Recognition (CVPR)}, pages 9556--9567.

\bibitem[{Zhou et~al.(2023)Zhou, Gao, Ding, Zheng, Xu, Wei, Zhang, and Ke}]{zhou2023unimol}
Gengmo Zhou, Zhifeng Gao, Qiankun Ding, Hang Zheng, Hongteng Xu, Zhewei Wei, Linfeng Zhang, and Guolin Ke. 2023.
\newblock Uni-mol: A universal 3d molecular representation learning framework.
\newblock In \emph{International Conference on Learning Representations (ICLR)}.

\end{thebibliography}

\appendix
% \newpage

\onecolumn
% \DoToC
% \clearpage

% \twocolumn

\setcounter{table}{0}

\renewcommand\thetable{\Alph{section}.\arabic{table}}

\setcounter{figure}{0}

\renewcommand\thefigure{\Alph{section}.\arabic{figure}}

\newpage

\section{Related Work}

Recent advancements in large language models (LLMs) have led to the development of sophisticated language agents capable of assisting in various aspects of chemical research \citep{ramos2024chemistryagentsurvey}. These agents, such as ChemCrow \citep{bran2023chemcrow} and Coscientist \citep{boiko2023coscientist}, have demonstrated the ability to automate routine chemical tasks and accelerate molecular discovery. ChemCrow, for instance, integrates LLMs with common chemical tools to perform a wide range of chemistry-related tasks, consistently outperforming GPT-4 \citep{achiam2023gpt4} in accuracy. Similarly, Coscientist exemplifies the integration of semi-autonomous robots in planning and executing chemical reactions with minimal human intervention. Other notable agents include Chemist-X \citep{chen2024chemistx}, which focuses on designing chemical reactions to achieve specific molecules, and ProtAgent \citep{ghafarollahi2024protagents}, a multi-agent system designed to automate and optimize protein design.
In the realm of experimental planning, several agents have been developed to bridge the gap between virtual assistants and physical laboratory environments. CALMS \citep{cherukara2024calms} enhances laboratory efficiency by operating instruments and managing complex experiments through conversational LLMs. BioPlanner \citep{o2023bioplanner} improves experimental efficiency by creating pseudocode representations of procedures, while CRISPR-GPT \citep{huang2024crispr} assists in designing gene editing experiments iteratively with constant human feedback. LLM-RDF \citep{ruan2024accelerated} takes this a step further by automating every step of the synthesis workflow, from literature search to product purification.
Cheminformatics tasks have also been significantly impacted by LLM-based agents. CACTUS \citep{mcnaughton2024cactus} automates the application of multiple cheminformatics tools while maintaining human oversight in molecular discovery. ChatMOF \citep{kang2023chatmof} focuses on predicting and generating Metal-Organic Frameworks, integrating MOF databases with its predictor module. IBM ChemChat augments LLMs with common APIs and Python packages used in cheminformatics research, facilitating tasks such as de novo drug design and property prediction. Most recently, \newcontent{\citet{tang2025chemagent} concurrently proposed an agent framework with a dynamic and self-updating memory mechanism to improve LLMs in solving general chemistry questions, and} \citet{chen2024scienceagentbench} examine agents' performance in data-driven scientific discovery via coding and achieve better performance with domain knowledge insertion and self-debug \citep{chen2024teaching}. These advancements collectively demonstrate the transformative potential of AI agents in chemical research, streamlining processes, enhancing efficiency, and accelerating scientific discovery.

Although these above agents have been proposed to tackle specific chemistry applications, there lacks a comprehensive evaluation on how tool-augmented agents perform on various chemistry problems. This study aims to address this issue and provide actionable insights to shed the light for future directions.

\section{Tool Set}
\label{app:tool_set}

\subsection{Tools}

The tool set contains 29 tools ranging from general tools, molecule tools, to reaction tools. This section introduces all the tools in detail.

\noindent \textbf{General tools}: Provide broad information retrieval, web searching, and computational.

\begin{itemize}[leftmargin=*]
    \item AiExpert: A general-purpose LLM prompted to answer any questions when other tools cannot handle. We use GPT-4o or Claude-3.5-Sonnet in our experiments, identical to the backbone models of \modelname.
    \item PythonREPL: Executes Python commands and allows for package installation.
    \item WebSearch: Searches the internet for both general and domain-specific information, providing concise summaries of relevant content. This involves an LLM-based search service\footnote{\url{https://tavily.com/}.} that uses LLMs to summarize the search result, providing more straightforward and organized results.
    \item WikipediaSearch: Searches Wikipedia and provides summaries of related content.
\end{itemize}

\noindent \textbf{Molecule tools}: Offer various analyses, predictions, and conversions related to chemical compounds and their properties.

\begin{itemize}[leftmargin=*]
    \item BBBPPredictor: Predicts the probability of a compound penetrating the blood-brain barrier using the Uni-Mol model \citep{zhou2023unimol}.
    \item CanonicalizeSMILES: Converts SMILES representation to its canonical form with RDKit \citep{rdkit}.
    \item CompareSMILES: Determines if two molecule SMILES representations are identical.
    \item CountMolAtoms: Counts the number and types of atoms in a molecule.
    \item FunctionalGroups: Identifies functional groups present in a molecule.
    \item GetMoleculePrice: Retrieves the cheapest available price for a purchasable molecule.
    \item HIVInhibitorPredictor: Predicts the probability of a compound inhibiting HIV replication using the Uni-Mol model.
    \item IUPAC2SMILES: Converts IUPAC names to SMILES representation by searching PubChem, ChemSpace, or using the neural network based STOUT model \citep{rajan2021stout}.
    \item LogDPredictor: Predicts the octanol/water distribution coefficient (logD) at pH 7.4 using the Uni-Mol model.
    \item MolSimilarity: Computes the Tanimoto similarity between two molecules.
    \item MoleculeCaptioner: Generates a textual description of a molecule using neural networks using the MolT5 model \citep{edwards2022translation}.
    \item MoleculeGenerator: Creates SMILES representations based on molecular descriptions using neural networks using the MolT5 model \citep{edwards2022translation}.
    \item Name2SMILES: Converts common molecule names to SMILES representation.
    \item PatentCheck: Verifies if a molecule is patented.
    \item PubchemSearchQA: Searches and retrieves molecule/compound information from PubChem, a comprehensive database of chemical molecules and their activities. Given the information of a molecule/compound (SMILES, IUPAC name, orcommon name) and a related question, it retrieves the corresponding document from PubChem, and applies an instructed LLM (GPT-4o in our experiments) to briefly answer the input questions. Instead of directly returning the whole document, which is typically very long, this QA design reduces the irrelevant information in the context, so as to avoid distractions and length limit violation for LLMs.
    
    \item SELFIES2SMILES: Converts SELFIES \citep{krenn2019selfies} to SMILES representation.
    \item SMILES2Formula: Derives the molecular formula from SMILES representation using fixed algorithm implemented with RDKit.
    \item SMILES2IUPAC: Converts SMILES representation to IUPAC name by searching PubChem, ChemSpace, or using the neural network based STOUT model.
    \item SMILES2SELFIES: Converts SMILES representation to SELFIES representation.
    \item SMILES2Weight: Calculates the molecular weight from SMILES representation.
    \item SideEffectPredictor: Predicts the probabilities of a compound causing various side effects across 20 different categories using the Uni-Mol model.
    \item SolubilityPredictor: Predicts the log solubility of a compound in mol/L using the Uni-Mol model.
    \item ToxicityPredictor: Predicts the probability of a compound being toxic using the Uni-Mol model.
\end{itemize}

\noindent \textbf{Reaction tools}: Predict products of chemical reactions and suggest potential reactants for synthesizing given products.

\begin{itemize}[leftmargin=*]
    \item ForwardSynthesis: Predicts the products of a chemical reaction based on given reactants and reagents using IBM RXN for Chemistry \footnote{\url{https://rxn.res.ibm.com/rxn}.}.
    \item Retrosynthesis: Conducts single-step retrosynthesis, suggesting potential reactants for a given product using IBM RXN for Chemistry.
\end{itemize}

\subsection{Improvements of Tool Set}

We build an improved tool set based on that of ChemCrow \citep{bran2023chemcrow}. The purposes are:

\begin{itemize}[leftmargin=*]
    \item \textbf{Addressing non-functional tools}: Some tools in ChemCrow were not well-maintained and failed to function properly, so we fixed their bugs and made them function again. For example, RXNPredict encountered a ``project\_id not set'' error due to implementation issues, and we fixed this issue, resulting in our ForwardSynthesis tool.
    \item \textbf{Improving tool usability}: Certain tools did not meet practical needs, so we modified them to make them more practical and robust. For instance, the original web search tool in ChemCrow has limitations, such as inflexible input handling and truncation of output when results are lengthy, often cutting off critical information. We resolved this by replacing it with an LLM-based search service, which supports flexible input and generates summarized outputs.
    \item \textbf{Filling gaps in functionality}: The original tool set lacked support for many important chemistry tasks, so we created tools to enable it on more diverse tasks. For example, accessing information from PubChem, a comprehensive database of molecules and compounds, is not supported. To address this, we implemented the PubChemQA tool. Chemical representation conversion is another critical capability missing from the original tool set. To fill this gap, we added multiple tools for the conversion.
\end{itemize}

After these improvements, the tool set became significantly more robust and comprehensive. This is reflected in the performance comparisons in \cref{tab:smolinstruct} and \cref{tab:mmluc_gpqac}, where \modelname consistently and substantially outperforms ChemCrow that uses the original tool set.

% While we tried our best to improve the tools, it is difficult to fully solve an intrinsic limitation – inaccurate or incorrect tool outputs. For example, some neural network-based tools, such as the molecular property predictors, may generate incorrect property values. Similarly, the web search tool can occasionally return inaccurate information due to the presence of unreliable or unauthorized sources on the internet. Future agent researchers should not only constantly optimize and maintain tools, but also enhance the agent’s ability to verify tool outputs automatically, such as through self-reflection or cross-verification mechanisms.

\section{Experiment Details}
\label{app:detail_sm}

\subsection{\smolinstruct}

\subsubsection{Dataset and Evaluation Setups}
\label{app:smolinstruct_tasks}

\begin{figure*}[thbp]
    \centering
    \includegraphics[width=0.9\textwidth]{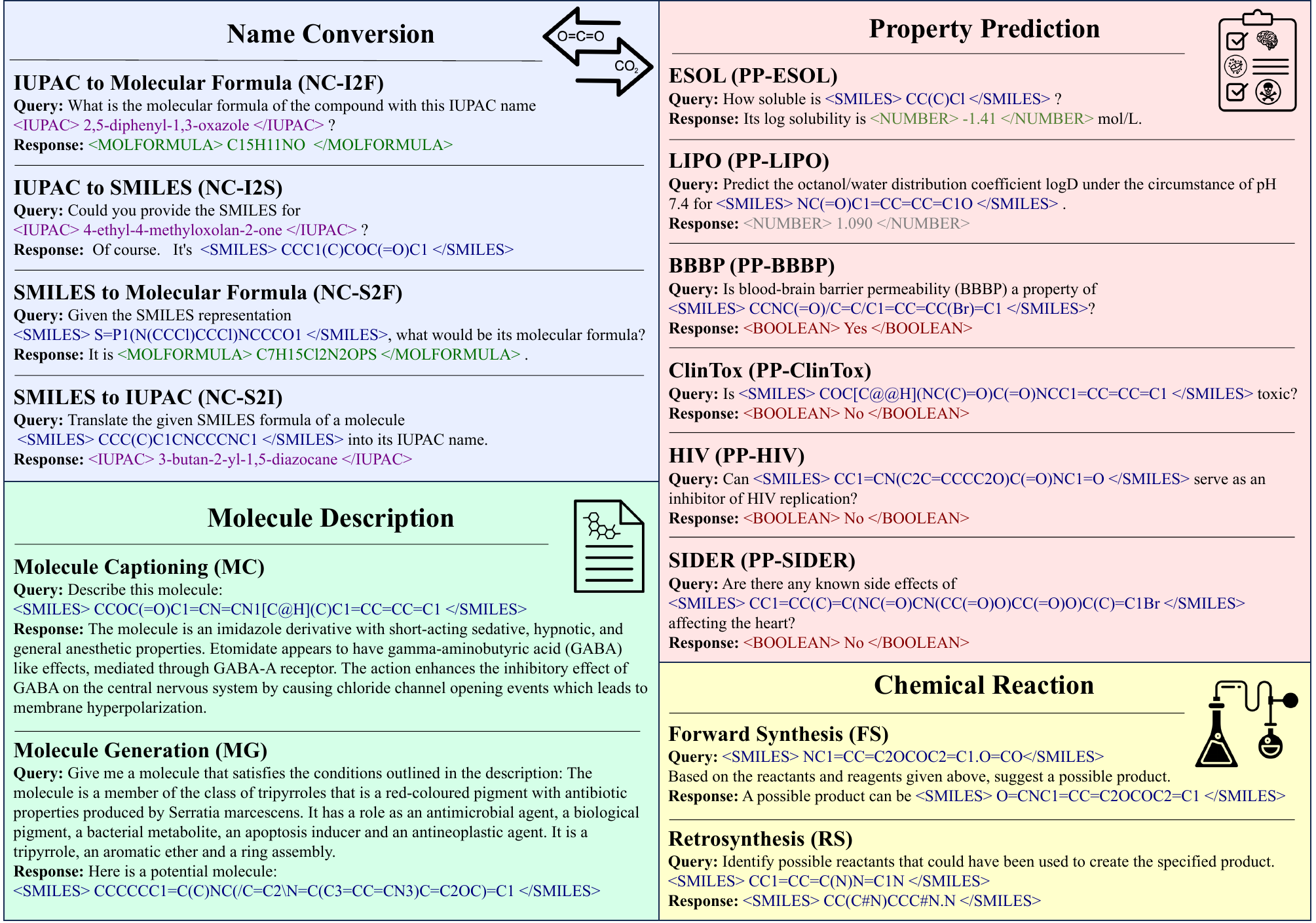}
    \caption{Tasks in SMolInstruct \citep{yu2024llasmol}.}
    \label{fig:smolinstruct_tasks}
    % }
\end{figure*}

\smolinstruct \citep{yu2024llasmol} contains 14 molecule- and reaction-centric tasks, which, along with the task name abbreviations and examples, are illustrated in \cref{fig:smolinstruct_tasks}.

We evaluate the models on 50 randomly selected samples from the test set for each task.
For reference, in the following detailed results, we also include the SoTA non-LLM models used in \citet{yu2024llasmol}, and LlaSMol\footnote{\href{https://huggingface.co/osunlp/LlaSMol-Mistral-7B}{osunlp/LlaSMol-Mistral-7B}.}, which is a Mistral model \citep{jiang2023mistral} fine-tuned on \smolinstruct.
For SoTA non-LLM models and LlaSMol, we adopt their own formats of input and output. For other models, we prompt them to think step by step, i.e., using chain-of-thought (CoT) \citep{wei2022chain,wang2022iteratively}, and wrap their final answers with ``\textless ANSWER\textgreater'' and ``\textless /ANSWER\textgreater'' to facilitate answer extraction.
The evaluation metrics are adopted from \citet{yu2024llasmol}.

\subsubsection{Detailed Results}
\label{app:detail_sm_results}

% Table generated by Excel2LaTeX from sheet 'Overall_SMolInstruct_1'
\begin{table*}[thbp]
  \centering
  % \caption{The results on \smolinstruct for name conversion and property prediction tasks. All the metrics except RMSE are in percentage.}
  \scalebox{0.7}{
    \sisetup{detect-weight=true}
    \begin{tabular}{lS[table-format=3.1]S[table-format=2.1]S[table-format=3.1]S[table-format=3.1]S[table-format=2.1]S[table-format=1.3]S[table-format=1.3]S[table-format=2.1]S[table-format=2.1]S[table-format=2.1]S[table-format=2.1]}
    \toprule
    \multicolumn{1}{c}{\multirow{3}[6]{*}{\textbf{Model}}} & \multicolumn{5}{c}{\textbf{NC}}       & \multicolumn{6}{c}{\textbf{PP}} \\
    \cmidrule(lr){2-6}  \cmidrule(lr){7-12}   & \textbf{I2F} & \multicolumn{2}{c}{\textbf{I2S}} & \textbf{S2F} & \textbf{S2I} & \textbf{ESOL} & \textbf{Lipo} & \textbf{BBBP} & \textbf{Clintox} & \textbf{HIV} & \textbf{SIDER} \\
     \cmidrule(lr){2-2} \cmidrule(lr){3-4} \cmidrule(lr){5-5} \cmidrule(lr){6-6} \cmidrule(lr){7-7} \cmidrule(lr){8-8} \cmidrule(lr){9-9} \cmidrule(lr){10-10}  \cmidrule(lr){11-11}  \cmidrule(lr){12-12}
     & \textbf{EM} & \textbf{EM} & \textbf{Valid} & \textbf{EM} & \textbf{EM} & \textbf{RMSE} $\downarrow$ & \textbf{RMSE} $\downarrow$ & \textbf{Acc} & \textbf{Acc} & \textbf{Acc} & \textbf{Acc} \\
    \midrule
    SoTA non-LLM models & 96.0  & \bfseries 68.0  & \bfseries 100.0 & \bfseries 100.0 & 54.0  & \bfseries 0.808 & \bfseries 0.527 & 88.0  & \bfseries 90.0  & \bfseries 94.0  & 70.0 \\
    GPT-4o & 12.0  & 0.0   & 66.0  & 8.0   & 0.0   & 1.315 & 1.264 & 70.0  & 36.0  & 86.0  & 44.0 \\
    Claude-3.5-Sonnet & 4.0   & 10.0  & 70.0  & 4.0   & 2.0   & 1.443 & 1.267 & 78.0  & 50.0  & 88.0  & 62.0 \\
    LlaSMol & 92.0  & 60.0  & 96.0  & 96.0  & 34.0  & 1.062 & 1.164 & 82.0  & 98.0  & \bfseries 94.0  & \bfseries 74.0 \\
    \midrule
    ChemCrow (GPT) & 18.0  & 10.0  & 18.0  & 88.0  & 2.0   & 4.376 & 2.061 & 46.0  & 62.0  & 74.0  & 36.0 \\
    ChemCrow (Claude) & 16.0  & 14.0  & 18.0  & 42.0  & 2.0   & 2.025 & 1.179 & 60.0  & 34.0  & 92.0  & 32.0 \\
    \midrule
    \modelsname (GPT) & \bfseries 100.0 & 64.0  & \bfseries 100.0 & \bfseries 100.0 & \bfseries 70.0  & 0.812 & 0.529 & \bfseries 90.0  & 82.0  & \bfseries 94.0  & 70.0 \\
    \modelsname (Claude) & \bfseries 100.0 & \bfseries 68.0  & \bfseries 100.0 & \bfseries 100.0 & \bfseries 70.0  & 1.131 & 0.531 & \bfseries 90.0  & 58.0  & 92.0  & 68.0 \\
    \bottomrule
    \end{tabular}%
    }
    \caption{The results on \smolinstruct for name conversion and property prediction tasks. All the metrics except RMSE are in percentage.}
  \label{tab:smolinstruct1}%
\end{table*}%

\vspace{-0.2cm}
% Table generated by Excel2LaTeX from sheet 'Overall_SMolInstruct_2'
\begin{table*}[thbp]
  \centering
  % \caption{The results on \smolinstruct for task MC, MG, FS, and RS. All the metrics except METEOR are in percentage.}
  \scalebox{0.8}{
    \sisetup{detect-weight=true}
    \begin{tabular}{lS[table-format=1.3]S[table-format=2.1]S[table-format=2.1]S[table-format=3.1]S[table-format=2.1]S[table-format=2.1]S[table-format=3.1]S[table-format=2.1]S[table-format=2.1]S[table-format=3.1]}
    \toprule
    \multicolumn{1}{l}{\multirow{2}[2]{*}{\textbf{Model}}} & \multicolumn{1}{c}{\textbf{MC}} & \multicolumn{3}{c}{\textbf{MG}} & \multicolumn{3}{c}{\textbf{FS}} & \multicolumn{3}{c}{\textbf{RS}} \\
    \cmidrule(lr){2-2} \cmidrule(lr){3-5} \cmidrule(lr){6-8} \cmidrule(lr){9-11}
          & \multicolumn{1}{c}{\textbf{METEOR}} & \multicolumn{1}{c}{\textbf{EM}} & \multicolumn{1}{c}{\textbf{FTS}} & \multicolumn{1}{c}{\textbf{Valid}} & \multicolumn{1}{c}{\textbf{EM}} & \multicolumn{1}{c}{\textbf{FTS}} & \multicolumn{1}{c}{\textbf{Valid}} & \multicolumn{1}{c}{\textbf{EM}} & \multicolumn{1}{c}{\textbf{FTS}} & \multicolumn{1}{c}{\textbf{Valid}} \\
    \midrule
    SoTA non-LLM models & \bfseries 0.539 & 32.0  & 75.7  & 96.0  & 78.0  & 91.7  & \bfseries 100.0 & \bfseries 42.0  & \bfseries 80.5  & \bfseries 100.0 \\
    GPT-4o & 0.152 & 10.0  & 57.5  & 84.0  & 12.0  & 46.3  & 84.0  & 0.0   & 36.0  & 84.0 \\
    Claude-3.5-Sonnet & 0.211 & 12.0  & 67.5  & 90.0  & 22.0  & 60.9  & 98.0  & 0.0   & 45.7  & 90.0 \\
    LlaSMol & 0.426 & 22.0  & 67.0  & 98.0  & 56.0  & 83.4  & \bfseries 100.0 & 26.0  & 70.3  & 100.0 \\
    \midrule
    ChemCrow (GPT) & 0.195 & 34.0  & 79.9  & 68.0  & 72.0  & \bfseries 92.5  & 92.0  & 8.0   & 49.0  & 74.0 \\
    ChemCrow (Claude) & 0.255 & 40.0  & 81.0  & 86.0  & 70.0  & 90.5  & 92.0  & 22.0  & 0.0   & 90.0 \\
    \midrule
    \modelsname (GPT) & 0.510 & 28.0  & 76.8  & 90.0  & 78.0  & 92.1  & 98.0  & \bfseries 42.0  & 78.0  & 98.0 \\
    \modelsname (Claude) & 0.443 & \bfseries 44.0  & \bfseries 83.5  & \bfseries 100.0 & \bfseries 80.0  & 92.2  & \bfseries 100.0 & \bfseries 42.0  & 78.6  & \bfseries 100.0 \\
    \bottomrule
    \end{tabular}%
    }
    \caption{The results on \smolinstruct for the MC, MG, FS, and RS tasks. All the metrics except METEOR score are in percentage.}
  \label{tab:smolinstruct2}%
\end{table*}%

The detailed results on \smolinstruct are presented in \cref{tab:smolinstruct1} and \cref{tab:smolinstruct2}. We can see that:
(1) The SoTA LLMs, GPT-4o and Claude-3.5-Sonnet, demonstrate relatively low performance across all evaluated tasks, which underscores the persistent challenges faced by general-purpose LLMs in specialized chemistry domains, particularly in handling molecular representations such as SMILES and executing specialized chemical operations.
(2) On all tasks, \modelname achieves the best performance or close, confirming the benefits of specialized tools for the \smolinstruct tasks.
(3) While Claude-3.5-Sonnet generally outperforms GPT-4o, their performance as \modelname backbones is comparable.
This parity can be attributed to the nature of the \smolinstruct tasks, which primarily require effective tool utilization rather than extensive knowledge or complex reasoning abilities inherent to the LLMs themselves. 
Both LLMs demonstrate proficiency as ``tool users,'' effectively leveraging the provided resources to address the given tasks.

\subsubsection{Potential Data Leakage via Tools}
To ensure a fair evaluation on SmolInstruct \citep{yu2024llasmol}, we account for potential data leakage in our experiments.

For tools based on our self-trained models, we mitigated data leakage by training exclusively on samples from the SmolInstruct training set. This applies to MoleculeCaptioner, MoleculeGenerator, SolubilityPredictor, LogDPredictor, BBBPPredictor, ToxicityPredictor, HIVInhibitorPredictor, and SideEffectPredictor. Since test examples were excluded from the training set, these tools cannot leak test data.

For ForwardSynthesis and Retrosynthesis, which rely on IBM’s APIs, determining the original training data of their backend models (Molecular Transformer \citep{schwaller2019molecular}) is challenging, making potential data leakage uncertain. However, the SmolInstruct paper retrained Molecular Transformer on the training set and reported test set performance consistent with \modelname’s results. This suggests that if data leakage exists, it is minimal; otherwise, \modelname’s performance would significantly surpass the SmolInstruct benchmarks.

\subsection{\mmluc}
\label{app:mmluc}

\subsubsection{Dataset and Evaluation Setups}

To effectively and efficiently evaluate the models, we build \mmluc, a subset of 70 chemistry question samples derived from the widely-used MMLU dataset \citep{hendryckstest2021mmlu}.
Specifically, to increase the difficulty and differentiation of the questions, while avoiding erroneous samples presented in the original MMLU, we select samples that appear in both MMLU-Pro \citep{wang2024mmlupro} and MMLU-Redux \citep{gema2024mmluredux}. These two datasets are verified versions of MMLU, and MMLU-Pro has extended the answer options from 4 to 10 to introduce more challenges. When the gold standard answers from both sources match, we utilize the 10 options from MMLU-Pro. In cases of discrepancies, we manually review and correct any potential issues. To reduce the cost of evaluation, we eliminated samples where all models performed correctly in our preliminary experiments. This results in a final set of 70 questions, divided evenly between 35 high school-level and 35 college-level questions.

In our evaluation, all the models are prompted to generate a CoT solution and close the solution with ``the answer is ...'' to facilitate answer extraction. To mitigate randomness, we run each sample three times and report the average accuracy.

In addition, to understand the influence of in-context examples, in the following detailed results, we also introduce a 5-shot setting in comparison with 0-shot for the base LLMs and \modelname. The questions of the in-context examples are originally from MMLU's and MMLU-Pro's development set, and we manually construct CoT solutions for the base LLMs and tool-using step-wise solutions for \modelname. The order of the examples is randomized for each test sample. 

\subsubsection{Detailed Results}
\label{app:detail_mmluc_results}

\begin{table}[htbp]
  \centering
  \sisetup{detect-weight=true}
  \resizebox{0.5\textwidth}{!}{
    \begin{tabular}{l
                    S[table-format=2.1]
                    S[table-format=2.1]
                    S[table-format=2.1]
                    }
    \toprule
    \textbf{Model} & \multicolumn{1}{c}{\textbf{High school}} & \multicolumn{1}{c}{\textbf{College}} & \multicolumn{1}{c}{\textbf{Overall}} \\
    \midrule
    GPT-4o (0-shot) & \bfseries 88.6 & 72.4 & 80.5 \\
    GPT-4o (5-shot) & 85.7 & 72.4 & 79.0  \\
    Claude-3.5-Sonnet (0-shot) & 83.8 & 69.5 & 76.7  \\
    Claude-3.5-Sonnet (5-shot) & 83.8 & 73.3 & 78.6 \\
    \midrule
    ChemCrow (GPT, 0-shot) & 47.6 & 39.0 & 43.3 \\
    ChemCrow (Claude, 0-shot) & 69.5 & 67.6 & 68.6 \\
    \midrule
    \modelsname (GPT, 0-shot) & 81.9 & 60.0 & 71.0 \\
    \modelsname (GPT, 5-shot) & 87.6 & 63.8 & 75.7 \\
    \modelsname (Claude, 0-shot) & 73.3 & 66.7 & 70.0 \\
    \modelsname (Claude, 5-shot) & 86.7 & \bfseries 79.0 & \bfseries 82.9 \\
    \bottomrule
    \end{tabular}%
    }
  \caption{Accuracies (\%) on \mmluc, averaged over three runs.}
  \label{tab:mmluc}%
\end{table}%

The results are presented in \cref{tab:mmluc}. 
(1) While \modelname achieves the highest overall performance in one specific configuration (Claude, 5-shot), it demonstrates inferior performance compared to the base LLMs in all other configurations. 
This trend persists across both high school and college questions and is also observed with ChemCrow, suggesting a consistent pattern rather than an isolated occurrence.
(2) Comparing 0-shot and 5-shot performance, the addition of examples (5-shot) yields minimal improvement for base LLMs but results in significant enhancement for \modelname. This disparity may be attributed to the extensive pre-training of base LLMs on general chemistry questions, potentially rendering additional examples redundant for task comprehension. Conversely, for \modelname, the step-wise demonstration examples appear to effectively guide the LLMs in reasoning and tool utilization and potentially reduce the cognitive overload for LLMs, thereby optimizing the problem-solving process. This finding suggests that incorporating examples can be a valuable strategy for enhancing the performance.

\subsection{\newcontent{\sbc}}
\label{app:detail_sbc_results}

\subsubsection{Dataset and Evaluation Setups}

\newcontent{We evaluate models using \sbc, a set of 223 chemistry questions from the SciBench dataset \citep{wang2024scibench}, designed to assess their proficiency in college-level chemistry knowledge and complex reasoning, particularly in calculations. For each question, if an answer unit is specified, we append ``The unit of your answer should be ...'' to the model input. All evaluated models are prompted to generate CoT solutions and conclude their output with ``the answer is ...'' to facilitate answer extraction. We report the average accuracy over three runs.}

\newcontent{In addition to the base LLMs and ChemCrow, we also compare our approach with the most recent concurrent work, ChemAgent \citep{tang2025chemagent}. To ensure a fair comparison, we made our best effort to evaluate their agent using our setup. However, due to certain issues in their codebase, this was not feasible. Therefore, we instead evaluate our agent using their setup and directly compare our performance with the results reported in their paper. Specifically, following their setup, we use gpt-4-1106-preview as the LLM backbone and the dataset version released in their repository (rather than the latest version available in the SciBench repository that we use for other experiments). The comparison results are shown in the bottom section of \cref{tab:scibench_detail}.}

\subsubsection{Detailed Results}

% Table generated by Excel2LaTeX from sheet 'SciBench_final'
\begin{table}[htbp]
  \centering
  \sisetup{detect-weight=true}
  \resizebox{0.6\textwidth}{!}{
    \begin{tabular}{l
                    S[table-format=2.1]
                    S[table-format=2.1]
                    S[table-format=2.1]
                    S[table-format=2.1]
                    S[table-format=2.1]
                    }
    \toprule
    \textbf{Model} & \textbf{atkins} & \textbf{chemmc} & \textbf{matter} & \textbf{quan} & \textbf{Overall} \\
    \midrule
    GPT-4o & 62.5  & 71.1  & 51.8  & 55.6  & 60.7 \\
    Claude-3.5-Sonnet & 62.5  & 77.2  & 50.4  & 50.5  & 60.7 \\
    \midrule
    ChemCrow (GPT-4o) & 58.4  & 60.5  & 34.0  & 50.5  & 52.5 \\
    ChemCrow (Claude) & 59.4  & 65.8  & 41.8  & 54.5  & 56.1 \\
    \midrule
    \modelsname (GPT-4o) & 63.8  & 64.9  & 56.0  & 48.5  & 60.1 \\
    \modelsname (Claude) & \bfseries 66.3  & \bfseries 79.8  & \bfseries 56.7  & \bfseries 61.6  & \bfseries 65.9 \\
    \midrule
    \midrule
    ChemAgent (GPT-4) & 61.2  & \bfseries 74.4  & 49.0  & 44.1  & 57.2 \\
    \modelsname (GPT-4) & \bfseries 65.1  & 70.1  & \bfseries 54.4  & \bfseries 45.1  & \bfseries 60.7 \\
    \bottomrule
    \end{tabular}%
    }
    \caption{\newcontent{Accuracies (\%) on \sbc, averaged over three runs. ``atkins'', ``chemmc'', ``matter'', and ``quan'' represent the four sources of chemistry questions in SciBench. The bottom section uses a different LLM backbone and dataset version for a fair comparison with ChemAgent \citep{tang2025chemagent}.}}
  \label{tab:scibench_detail}%
\end{table}%

\newcontent{When compared to the base models and ChemCrow, \modelsname (Claude) achieves the highest accuracies. This is largely due to its frequent use of the PythonREPL tool for calculations, which is particularly beneficial for numerical questions in this dataset. However, \modelsname (GPT) still slightly underperforms its base model, GPT-4o, suggesting that while access to multiple tools can be advantageous, the added complexity and potential errors introduced by tool use may offset these benefits.}

\newcontent{In comparison to ChemAgent \citep{tang2025chemagent} (as shown in the last section of \cref{tab:scibench_detail}), \modelsname demonstrates stronger performance. This suggests that while ChemAgent’s sophisticated memory mechanism may offer benefits in certain situations, our straightforward method \modelsname proves to be very effective.}

\subsection{\gpqac}

We use \gpqac, the 93 chemistry multi-choice questions from the expert-verified GPQA-Diamond subset of the GPQA dataset \citep{rein2023gpqa} to evaluate models' abilities in high-level chemistry knowledge and complex reasoning.
All the evaluated models are prompted to generate CoT solutions and close their output with ``the answer is ...'' to facilitate answer extraction. We report the average accuracy across 3 runs. The results are presented in \cref{tab:mmluc_gpqac}.

\section{Error Cases}
\label{app:errors}

This section presents a specific case of each error. For reasoning errors, the cases are shown in \cref{ec:reasoning_wrong_knowledge_reasoning_incomplete_reasoning}, \cref{ec:reasoning_wrong_final_answer}, \cref{ec:reasoning_info_oversight}, and \cref{ec:reasoning_algebra_error}. For grounding errors, the case is presented in \cref{ec:grounding_wrong_input_format}. For tool errors, the cases are presented in \cref{ec:tool_wrong_tool_output} and \cref{ec:tool_inconsistent_tool_outputs}.

\subsection{Reasoning: Wrong Knowledge/Logic and Incomplete Reasoning}
\label{ec:reasoning_wrong_knowledge_reasoning_incomplete_reasoning}

\noindent \textbf{Explanation}

{\noindent\small This is an example of two reasoning errors: \uline{wrong knowledge/logic} and \uline{incomplete reasoning}. The question is from \mmluc, and \modelname answers it by firstly calculating the amount of the reactants in moles with the PythonREPL tool and then analyzing it based on its internal knowledge. The calculation with PythonREPL is correct, but the \hl{highlighted} conclusion is incorrect. While the agent concludes that the limiting reactant is bromine, the rubidium is actually the limiting reactant in this case. Additionally, the model does not verify whether the amounts of the remaining reactants and the product are larger than 0.10 mg, showcasing an incomplete reasoning.}

\noindent \textbf{Question}

{\noindent\small If 1.0 g of rubidium and 1.0 g of bromine are reacted, what will be left in measurable amounts (more than 0.10 mg) in the reaction vessel?} \\
{\noindent\small (A) RbBr only \quad
(B) RbBr, Rb, Br2, and Rb2Br \quad
(C) RbBr and Rb2Br only} \\
{\noindent\small (D) RbBr, Rb, and Br2 \quad
(E) Rb and Br2 only \quad
(F) Br2 only }\\
{\noindent\small (G) RbBr and Rb only \textbf{[predicted]} \quad
(H) Rb only} \\
{\noindent\small (I) Nothing will be left in measurable amounts \quad
(J) RbBr and Br2 only \textbf{[Correct]}}\\

\noindent {\textbf{Problem Solving Process of \modelname (GPT)}}

\steptt{1}\\
    \sctt{Thought}
    \begin{shaded*}\vspace{-0.5em}
    {\noindent\small 
    To determine what substances will be left in measurable amounts, I need to conduct a stoichiometric calculation based on the reaction between rubidium (Rb) and bromine (Br2) to form rubidium bromide (RbBr). First, I'll calculate the moles of rubidium and bromine to verify which one is the limiting reagent. Then I will determine whether any excess reactant is left after the reaction.
    Given:
    \begin{itemize}[leftmargin=*]
        \item Atomic weight of Rubidium (Rb) = 85.47 g/mol 
        \item Molecular weight of Bromine (Br2) = 159.808 g/mol (since Br = 79.904 g/mol)
    \end{itemize}
\small Let's start by calculating the moles of rubidium and Bromine.}
\vspace{-0.5em}\end{shaded*}
    
    \sctt{Action} 
    \begin{shaded*}\vspace{-0.5em}
    {\noindent\small Tool: PythonREPL
    
    \noindent Tool input:
    
    \begin{lstlisting}[language=Python, xleftmargin=0.3cm]
rubidium_molar_mass = 85.47  # g/mol for Rubidium (Rb)
bromine_molar_mass = 159.808  # g/mol for Bromine (Br2)
# Calculate moles
rubidium_moles = 1.0 / rubidium_molar_mass
bromine_moles = 1.0 / bromine_molar_mass
rubidium_moles, bromine_moles\end{lstlisting}}\vspace{-0.5em}\end{shaded*}
    
    \sctt{Observation}
    \begin{shaded*}\vspace{-0.5em}
    \noindent \small
        Tool output: (0.0117000117000117, 0.006257509010812976)
    \vspace{-0.5em}\end{shaded*}

    \steptt{2}\\
    \sctt{Thought}
    \begin{shaded*}\vspace{-0.5em}
    \noindent\small
    The moles of rubidium are approximately 0.0117 mol, and the moles of bromine are approximately 0.0063 mol. The reaction between rubidium and bromine to form rubidium bromide (RbBr) is as follows:
    $ 2 \text{Rb} + \text{Br}_2 \rightarrow 2 \text{RbBr} $

    \noindent From the stoichiometry of the reaction, 2 moles of rubidium react with 1 mole of bromine. Therefore, \hl{bromine is the limiting reagent} since 0.0117 moles of rubidium would require 0.00585 moles of bromine (less than 0.0063 moles provided).
    Since bromine is the limiting reagent, it will be completely consumed, while excess rubidium will remain.
    
    \noindent Now I can determine what substances will be present in measurable amounts ($>$ 0.10 mg) based on the stoichiometric calculation: 
    
    \noindent 1. Bromine will be fully consumed.
    
    \noindent 2. Rubidium bromide (RbBr) will be formed.
    
    \noindent 3. Excess rubidium will remain unreacted.
    
    \noindent Therefore, RbBr and excess Rb will be present in measurable amounts.\vspace{-0.5em}\end{shaded*}

    \sctt{Answer}
    
    \begin{shaded*}
    \noindent \small The answer is (G) RbBr and Rb only.
    \end{shaded*}

\subsection{Reasoning: Wrong Final Answer}
\label{ec:reasoning_wrong_final_answer}

\noindent \textbf{Explanation}

{\noindent \small This is an example of a \uline{reasoning error: wrong final answer}. The question is from \mmluc, and the agent answers it with the LLM's internal knowledge and no tool. Although it clearly rules out option (B) as \hl{highlighted} in its analysis, it still selects (B) as the final answer, which is wrong.}
    
\noindent \textbf{Question}

{\noindent \small
Which of the following is a true statement about optical isomerism of complexes containing achiral ligands?\\
(A) Square planar complexes can display optical isomerism only if all four ligands are identical.\\
(B) Tetrahedral complexes never display optical isomerism. \textbf{[predicted]}\\
(C) Linear complexes can display optical isomerism when both ligands are different.\\
(D) Octahedral complexes of monodentate ligands can display optical isomerism only when they have at least three different ligands. \textbf{[correct]}}

\noindent {\textbf{Problem Solving Process of \modelname (GPT)}}

    \steptt{1}\\
    \sctt{Thought}
    \begin{shaded*}
    
    \small{
    \noindent To answer this question, we need to understand the coordination geometries and when optical isomerism can occur in complexes. Let's briefly consider each statement:

\noindent (A) Square planar complexes: Optical isomerism occurs when there is no plane of symmetry in the molecule. Square planar complexes typically don't show optical isomerism unless they're part of a larger chiral structure or have different ligands.

\noindent (B) Tetrahedral complexes: \hl{Tetrahedral complexes can exhibit optical isomerism} if they have four different ligands, forming a chiral center analogous to asymmetric carbon in organic chemistry.

\noindent (C) Linear complexes: Linear complexes generally don't exhibit optical isomerism, as they require stereocenters which are not possible with only two ligands.

\noindent (D) Octahedral complexes: Octahedral complexes can show optical isomerism when they have a specific arrangement of different ligands that break symmetry and create chirality.

\noindent The only true statement regarding optical isomerism in complexes that could contain achiral ligands is (B).}
\end{shaded*}

    \sctt{Answer}
    
    \begin{shaded*}
    
    \noindent \small The answer is (B).
    
    \end{shaded*}

\subsection{Reasoning: Information Oversight}
\label{ec:reasoning_info_oversight}

\noindent \textbf{Explanation}

{\noindent \small This is an example of a \uline{reasoning error: information oversight}. The question is from \mmluc, and the agent answers it by using PythonREPL to calculate the result. However, the model ignores ``13C'' in the question, which indicates that the frequency should be approximately 1/4 of the frequency of the NMR instrument. This oversight leads to the wrong final answer.}
    
\noindent \textbf{Question}

{\noindent \small
The 13C chemical shifts of benzene and chloroform are 128.4 ppm and 77.2 ppm respectively. What is the difference in the 13C NMR frequencies of the two compounds on a 600 MHz spectrometer?\\
(A) 7.73 kHz \textbf{[correct]} \quad
(B) 60.3 kHz \quad
(C) 122 kHz \quad
(D) 500 kHz \quad
(E) 15.4 kHz \\
(F) 30.7 kHz \textbf{[predicted]} \quad
(G) 183.2 kHz \quad
(H) 367.2 kHz \quad
(I) 91.6 kHz \quad
(J) 244 kHz
}

\noindent {\textbf{Problem Solving Process of \modelname (GPT)}}

    \steptt{1}\\
    \sctt{Thought}
    \begin{shaded*}\noindent  \small
    To find the difference in frequencies, we need to calculate the frequency difference corresponding to the chemical shift difference. This can be calculated using the formula: frequency difference (in Hz) = chemical shift difference (in ppm) × spectrometer frequency (in MHz).

    \noindent The chemical shift difference between benzene and chloroform is (128.4 ppm - 77.2 ppm), and the spectrometer frequency is 600 MHz. Let's calculate the frequency difference in kHz.\end{shaded*}

    \sctt{Action}
    \begin{shaded*} \small
    \noindent Tool: PythonREPL
    
    \noindent Tool input:
    \begin{lstlisting}[language=Python, xleftmargin=0.3cm]
chemical_shift_difference = 128.4 - 77.2
spectrometer_frequency_mhz = 600
# Calculating the frequency difference in kHz
frequency_difference_khz = chemical_shift_difference * spectrometer_frequency_mhz
frequency_difference_khz
\end{lstlisting}\end{shaded*}

    \sctt{Observation}
    \begin{shaded*} \noindent \small
    Tool output: 30720.0
    \end{shaded*}

    \steptt{2} \\
    \sctt{Thought}
    \begin{shaded*} \noindent \small
    The frequency difference between benzene and chloroform on a 600 MHz spectrometer is 30.72 kHz. This corresponds to option (F) 30.7 kHz.
    \end{shaded*}
    \vspace{-0.5em}
    \sctt{Answer}
    \begin{shaded*} \noindent \small
    The answer is (F).
    \end{shaded*}

\subsection{Reasoning: Algebra Error}
\label{ec:reasoning_algebra_error}

\noindent {\textbf{Explanation}}

{\noindent \small This is an example of a \uline{reasoning error: algebra error}. The question is from \mmluc, and the agent answers it by firstly using Name2SMILES to understand the molecular structures and then calculating the ratio on its own. Everything is correct until the \hl{highlighted} algebra calculation step fails to correctly cross-multiply to solve for the ratio, leading to the wrong final answer.}

\noindent {\textbf{Question}}

{\noindent \small
        The 1H spectrum of a mixture of dimethylsulphoxide (DMSO) and acetonitrile (AN) contains lines with relative intensities $\alpha$ and $3\alpha$, respectively. What is the ratio of the two concentrations, [DMSO]:[AN]?\\
(A) 3:2 \quad
(B) 1:6 \textbf{[correct]} \quad
(C) 1:9 \textbf{[predicted]} \quad
(D) 1:1 \quad
(E) 2:3 \\
(F) 1:3 \quad
(G) 3:1 \quad
(H) 2:1 \quad
(I) 6:1 \quad
(J) 1:2}\\

\vspace{-0.5em}
\noindent {\textbf{Problem Solving Process of \modelname (GPT)}}
    
    \steptt{1}\\
    \sctt{Thought}
    \begin{shaded*} \noindent \small
    The relative intensities in the 1H NMR spectrum of the mixture of DMSO and acetonitrile are given as $\alpha$ and $3\alpha$, respectively. To find the concentration ratio [DMSO]:[AN], the relative number of nuclei contributing to these intensities must be considered. Both DMSO and acetonitrile contribute distinct numbers of protons. I will determine the number of protons contributing to the NMR spectrum for each compound.\end{shaded*}

    \sctt{Action}
    \begin{shaded*} \noindent \small
    Tool: Name2SMILES\\
    Tool input: Dimethylsulphoxide
    \end{shaded*}

    \sctt{Observation}
    \begin{shaded*} \noindent \small
    Tool output: \texttt{CS(=O)C}
    \end{shaded*}

    \steptt{2} \\
    \sctt{Thought}
    \begin{shaded*} \noindent \small
    Dimethylsulphoxide (DMSO) has the SMILES representation \texttt{CS(=O)C}, meaning it has 6 equivalent hydrogen atoms (as CH3 groups contribute their protons). Now, I need to find out the SMILES for acetonitrile (AN) to determine its number of hydrogens.
    \end{shaded*}

    \sctt{Action}
    \begin{shaded*} \noindent \small
    Tool: Name2SMILES\\
    Tool input: Acetonitrile
    \end{shaded*}

    \sctt{Observation}
    \begin{shaded*} \noindent \small
    Tool output: \texttt{CC\#N}
    \end{shaded*}

    \steptt{3} \\
    \sctt{Thought}
    \begin{shaded*} \noindent \small
    Acetonitrile (AN) has the SMILES representation \texttt{CC\#N}, meaning it has 3 equivalent hydrogen atoms (from the CH3 group). 

    \noindent Dimethyl sulfoxide (DMSO) thus contributes 6 protons to the NMR, whereas acetonitrile (AN) contributes 3 protons. The NMR intensity is proportional to the number of protons, thus the ratio of concentrations can be calculated considering this proportionality.
    \begin{itemize}[leftmargin=*]
        \item The intensity ratio given in the NMR spectrum is $\alpha$ (for DMSO) and $3\alpha$ (for AN).
        \item Let the concentration of DMSO be [DMSO] and that of AN be [AN].
    \end{itemize}
    
    \noindent Based on proportionality:
    
\[ \frac{\text{[DMSO]}\times6}{\text{[AN]}\times3} = \frac{\alpha}{3\alpha}  \]
\begin{equation}
    \colorbox{yellow}{  $\displaystyle \frac{\text{[DMSO]}}{\text{[AN]}} = \frac{1}{9}  $   }
\end{equation}

\noindent The concentration ratio of DMSO to AN is therefore 1:9.\end{shaded*}

    \sctt{Answer}
    \begin{shaded*} \noindent \small
    The answer is (C) 1:9.
    \end{shaded*}

\subsection{Grounding: Wrong Input Format}
\label{ec:grounding_wrong_input_format}

\noindent \textbf{Explanation}

{\noindent \small This is an example of a \uline{grounding error: wrong input format}. This question is from \smolinstruct, and the agent initially calls the dedicated tool, SideEffectPredictor, to predict the potential side effect. However, it inputs an extra ``SMILES:'' (\hl{highlighted}), causing the tool to malfunction. The same issue occurs with the CanonicalizeSMILES tool.}

\noindent \textbf{Question}

{\noindent \small Are there any known side effects of \textless SMILES\textgreater \texttt{CC(C)(C)NC[C@H](O)COC1=NSN=C1N1CCOCC1.CCN[C@H]1CN(CCCOC)S(=O) (=O)C2=C1C=C(S(N)(=O)=O)S2} \mbox{\textless/SMILES\textgreater} affecting the hepatobiliary system? \\
\textbf{Correct:} No. \quad \textbf{Predicted:} Yes. \\}

\noindent {\textbf{Problem Solving Process of \modelname (GPT)}}

    \steptt{1}\\
    \sctt{Thought}

    \begin{shaded*} \noindent \small
    To determine if there are any known side effects of the given compound affecting the hepatobiliary system, I will first use the SideEffectPredictor tool to obtain the probabilities of different side effects caused by the compound. After that, I will specifically look for any effects related to the hepatobiliary system.
    \end{shaded*}

    \sctt{Action}
    \begin{shaded*} \noindent \small
    Tool: SideEffectPredictor
    
    \noindent Tool input: \hl{SMILES:} \texttt{CC(C)(C)NC[C@H](O)COC1=NSN=C1N1CCOCC1.CCN[C@H]1CN(CCCOC)S(=O)(=O)C2=C1C=C(S(N) (=O)=O)S2}
    \end{shaded*}

    \sctt{Observation}
    \begin{shaded*} \noindent \small
    Tool output: Error: Invalid SMILES.
    \end{shaded*}

    \steptt{2} \\
    \sctt{Thought}
    \begin{shaded*} \noindent \small
    The error indicates that the provided SMILES is invalid. I'll canonicalize the SMILES and then re-attempt using the SideEffectPredictor tool.
    \end{shaded*}

    \sctt{Action}
    \begin{shaded*} \noindent \small
    Tool: CanonicalizeSMILES
    
    \noindent Tool input: \hl{SMILES:} \texttt{CC(C)(C)NC[C@H](O)COC1=NSN=C1N1CCOCC1.CCN[C@H]1CN(CCCOC)S(=O)(=O)C2=C1C=C(S(N) (=O)=O)S2}
    \end{shaded*}

    \sctt{Observation}
    \begin{shaded*} \noindent \small
    Tool output: Error: Invalid SMILES.
    \end{shaded*}

    \steptt{2} \\
    \sctt{Thought}
    \begin{shaded*} \noindent \small
    Since there's an issue with the SMILES provided and attempts to canonicalize it failed, I'll use the AiExpert tool to get an analysis regarding the potential side effects based on the description of the molecule.
    \end{shaded*}

    \sctt{Action}
    \begin{shaded*} \noindent \small
    Tool: AiExpert
    
    \noindent Tool input: Are there any known side effects affecting the hepatobiliary system for a compound with both isothiazolone and sulfonamide-like structures?
    \end{shaded*}

    \sctt{Observation}
    \begin{shaded*} \noindent \small
    Tool output: Compounds that contain isothiazolone and sulfonamide-like structures can be complex in nature ... (omit the long analysis)
    \\Given the potential for systemic toxicity, especially with ongoing exposure, caution is warranted, and hepatobiliary monitoring should be part of safety assessments for such a compound. Always consult and follow established guidelines and protocols for safety evaluation during drug development or chemical utilization phases.
    \end{shaded*}

    \steptt{3} \\
    \sctt{Thought}
    \begin{shaded*} \noindent \small
    The AiExpert tool suggests there could be potential hepatobiliary side effects for compounds containing isothiazolone and sulfonamide-like structures due to possible hepatotoxicity, liver enzyme changes, or hypersensitivity reactions involving the liver. Therefore, there's a likelihood of such effects occurring in the given compound.
    \end{shaded*}
    \vspace{-0.5em}
    \sctt{Answer}
    \begin{shaded*} \noindent \small
    Yes.
    \end{shaded*}

\subsection{Tool: Wrong Tool Output}
\label{ec:tool_wrong_tool_output}

\noindent \textbf{Explanation}

{\noindent \small This example shows the \uline{tool: wrong tool output} error. The question is from \mmluc, and the agent answers it by searching the internet with the WebSearch tool. The tool provides an incorrect result (\hl{highlighted}), which leads to the wrong final answer.}

\noindent \textbf{Question}
        
{\noindent \small Of the following atoms, which has the lowest electron affinity? \\
(A) F \quad
(B) Si \textbf{[predicted]} \quad
(C) O \quad
(D) Ca \textbf{[gold]}}

\noindent {\textbf{Problem Solving Process of \modelname (GPT)}}

    \steptt{1}\\
    \sctt{Thought}
    \begin{shaded*} \noindent \small
    Electron affinity generally increases across a period and decreases down a group in the periodic table. Considering this trend, I need to examine the electron affinities of the given atoms: F, Si, O, and Ca. I will search for electron affinities to accurately determine which atom has the lowest value.
    \end{shaded*}

    \sctt{Action}
    \begin{shaded*} \noindent \small
    Tool: WebSearch
    \\
    \noindent Tool input: Which atom has the lowest electron affinity among F, Si, O, and Ca?
    \end{shaded*}

    \sctt{Observation}
    \begin{shaded*} \noindent \small
    Tool output: \hl{The atom with the lowest electron affinity among F, Si, O, and Ca is Silicon (Si).} It has an electron affinity value of 134.06 kJ/mol, which is lower compared to the other atoms mentioned.
    \end{shaded*}
    \vspace{-0.5em}
    \steptt{2} \\
    \sctt{Thought}
    \begin{shaded*} \noindent \small
    Based on the electron affinity values obtained, Silicon (Si) has the lowest electron affinity among the listed options.
    \end{shaded*}

    \sctt{Answer}
    \begin{shaded*} \noindent \small
    The answer is (B) Si.
    \end{shaded*}

\subsection{Tool: Inconsistent Tool Outputs}
\label{ec:tool_inconsistent_tool_outputs}

\noindent \textbf{Explanation}

{\noindent \small This is an example of a \uline{tool error: inconsistent tool outputs}. The question is from \smolinstruct, and the agent answers it by predicting the answer with ToxicityPredictor and verifying the answer with PubchemSearchQA. However, the two tools give different answers and the agent chooses to believe the latter, which leads to the wrong final answer.}

\noindent \textbf{Question}
        
{\noindent \small Is \textless SMILES\textgreater \texttt{C[C@H](/C=C/[C@H](C)C(C)(C)O[C@H]1CC[C@H]2/C(=C/C=C3C[C@@H](O)C[C@H](O)C3)CCC[C@]12C} \mbox{\textless/SMILES\textgreater} classified as a hazardous substance with documented toxicity?\\
\textbf{Correct:} No. \quad
\textbf{Predicted:} Yes.}

\noindent {\textbf{Problem Solving Process of \modelname (GPT)}}

    \steptt{1}\\
    \sctt{Thought}
    \begin{shaded*} \noindent \small
    To determine if the given SMILES represents a hazardous substance with documented toxicity, I should first predict its toxicity profile using the ToxicityPredictor tool.
    \end{shaded*}

    \sctt{Action}
    \begin{shaded*} \noindent \small
    Tool: ToxicityPredictor
    
    \noindent Tool input: \texttt{C[C@H](/C=C/[C@H](C)C(C)(C)O)[C@H]1CC[C@H]2/C(=C/C=C3C[C@@H](O) C[C@H](O)C3)CCC[C@]12C}
    \end{shaded*}

    \sctt{Observation}
    \begin{shaded*} \noindent \small
    Tool output: The probability of the compound to be toxic is 7.85\%, which means it's unlikely to happen. Note that the result is predicted by a neural network model and may not be accurate. You may use other tools or resources to obtain more reliable results if needed.
    \end{shaded*}

    \steptt{2} \\
    \sctt{Thought} 
    \begin{shaded*} \noindent \small
    The ToxicityPredictor tool suggests that the probability of this compound being toxic is relatively low (7.85\%), indicating it is unlikely to be classified as a hazardous substance based on this prediction alone. However, considering the neural network model's limitations, it would be prudent to verify this against more authoritative data sources like PubChem for documented toxicity.
    \end{shaded*}

    \sctt{Action}
    \begin{shaded*} \noindent \small
    Tool: PubchemSearchQA
    
    \noindent Tool input: SMILES: \texttt{C[C@H](/C=C/[C@H](C)C(C)(C)O)[C@H]1CC[C@H]2/C(=C/C=C3C[C@@H](O)C[C@H](O)C3)CCC [C@]12C}\\Question: Is this compound classified as a hazardous substance with documented toxicity?
    \end{shaded*}

    \sctt{Observation}
    \begin{shaded*} \noindent \small
    Tool output: Yes, paricalcitol is classified as a hazardous substance with documented toxicity. It has several GHS hazard classifications, including being fatal if swallowed (H300) and toxic if swallowed or in contact with skin (H301+H311). Additionally, it is suspected of causing cancer (H351) and may cause harm to breast-fed children (H362). Paricalcitol has acute toxicity both orally and dermally, and is recognized for causing damage to organs through prolonged or repeated exposure (H372).\\Furthermore, toxicological information indicates potential drug-induced liver injury (ambiguous concern), and toxicity symptoms such as hypercalcemia can be severe. It also shows potential for interactions that could exacerbate its toxic effects. These factors collectively contribute to its classification as a hazardous substance with significant documented toxicity.
    \end{shaded*}

    \steptt{3} \\
    \sctt{Thought}
    \begin{shaded*} \noindent \small
    Based on the information obtained from PubChem, the compound represented by the given SMILES, identified as paricalcitol, is indeed classified as a hazardous substance with documented toxicity. It has several GHS hazard classifications and potential health risks.
    \end{shaded*}

    \sctt{Answer}
    \begin{shaded*} \noindent \small
    Yes.
    \end{shaded*}

\clearpage
\section{Tool Usage Analysis}
\label{app:tool_analysis}

\begin{figure}[htbp]
    \centering
    \includegraphics[width=\columnwidth]{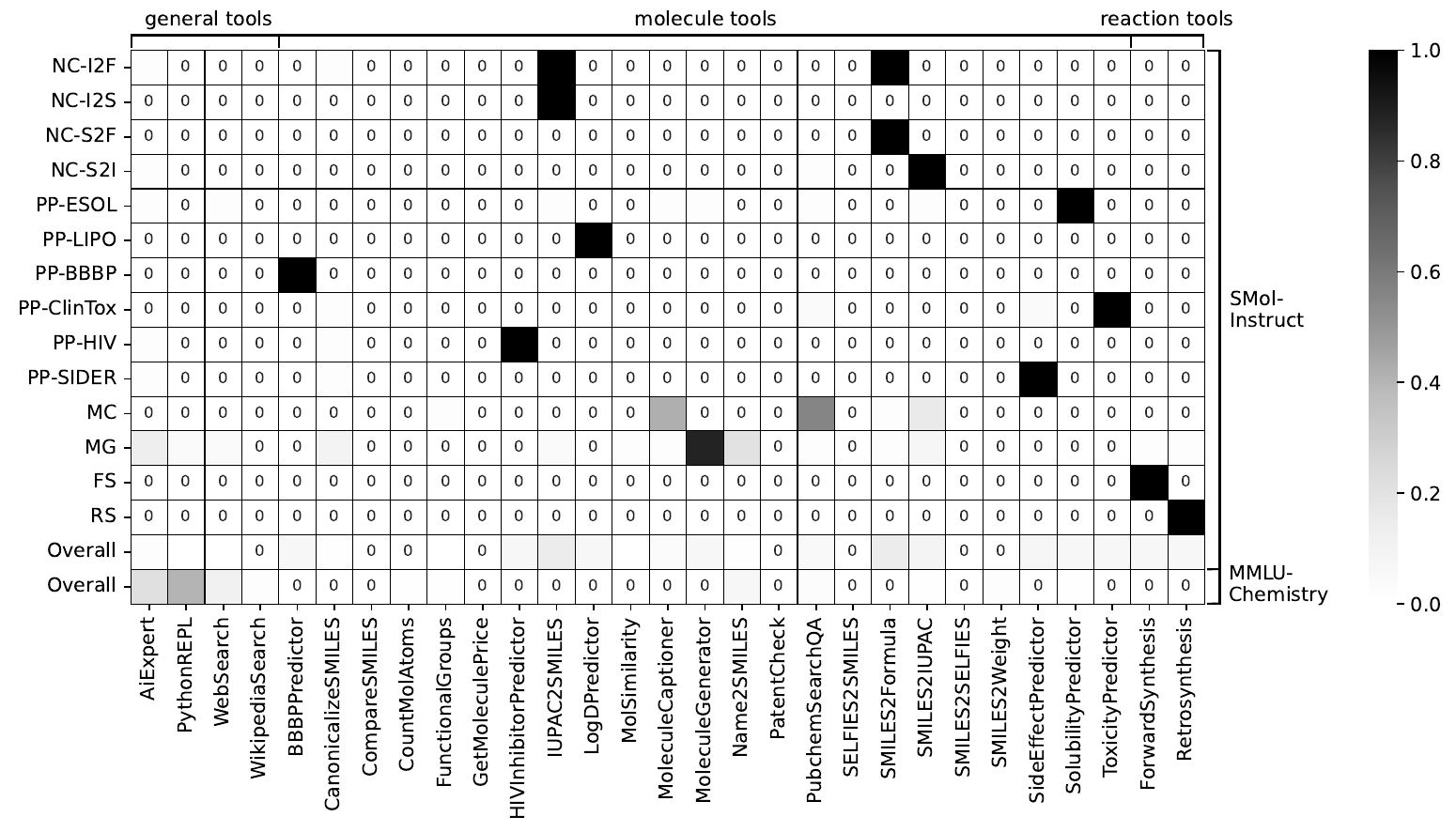}
    \caption{The statistics of tool usage by \modelname (GPT). The cell values represent the ratios of number of samples where the corresponding tools are used out of the total number of samples, and the number ``0'' indicates the value is 0 (the tool is not used).}
    \label{fig:tool_heatmap}
    % }
\end{figure}

To examine the tool usage by the agent, we perform a statistical analysis of the tools used on both \smolinstruct and \mmluc, as depicted in \cref{fig:tool_heatmap}. 

On \smolinstruct, the agent predominantly invokes the tools designed for the tasks. For instance, in the NC-I2F task, the agent commonly utilizes IUPAC2SMILES and SMILES2Formula to facilitate the conversion from IUPAC names to molecular formulas, with these tools often achieving values near 1.0. The molecular captioning (MC) and molecular generation (MG) tasks are notable exceptions. Since they are more open-ended, the agent opts for a variety of tools. 

Conversely, on \mmluc, the agent typically resorts to general-purpose tools (e.g., PythonREPL for calculations, WebSearch for knowledge gathering), due to the nature of the questions and the absence of task-specific tools.
% \newcontent{This indicates the need for balancing tool augmentation with internal reasoning capabilities for different tasks.}

\end{document}